\pdfoutput=1
\documentclass[11pt]{article}

\usepackage[]{EMNLP2023}
\usepackage{times}
\usepackage{latexsym}

\usepackage[T1]{fontenc}
\usepackage[utf8]{inputenc}

\usepackage{microtype}

\usepackage{inconsolata}

\usepackage{subfigure}
\usepackage{graphicx}
\usepackage{subfigure}
\usepackage{multirow}
\usepackage{makecell}
\usepackage{booktabs}
\usepackage{amsmath}
\usepackage{bm}
\usepackage{etoolbox}
\usepackage{amssymb}

\usepackage{hyperref}
\usepackage{color}
\usepackage{stfloats}

\title{AnnoLLM: Making Large Language Models to Be Better\\ Crowdsourced  Annotators}

\author{%
  Xingwei He\textsuperscript{\rm 1}\thanks{\ \ Work done during internship at Microsoft Research Asia.}, \quad
  Zhenghao Lin\textsuperscript{\rm 2}, \quad
  Yeyun Gong\textsuperscript{\rm 4}, \quad
  A-Long Jin\textsuperscript{\rm 3},  \quad
  Hang Zhang\textsuperscript{\rm 4}, 
  \\
  \textbf{Chen Lin}\textsuperscript{\rm 2}\textbf{,}  \quad
  \textbf{Jian Jiao}\textsuperscript{\rm 5}\textbf{,}  \quad 
  \textbf{Siu-Ming Yiu}\textsuperscript{\rm 1}\thanks{\ \ Corresponding author.}\textbf{,} \quad
  \textbf{Nan Duan}\textsuperscript{\rm 4}\textbf{,} \quad
  \textbf{Weizhu Chen}\textsuperscript{\rm 5}
  \\
  \textsuperscript{\rm 1}The University of Hong Kong,
  \textsuperscript{\rm 2}Xiamen University,\\
  \textsuperscript{\rm 3}Xi'an Jiaotong-Liverpool University,
 \textsuperscript{\rm 4}Microsoft Research Asia,
  \textsuperscript{\rm 5}Microsoft 
  \\
  \texttt{hexingwei15@gmail.com},  
  \texttt{along.jin@xjtlu.edu.cn}, 
  \texttt{smyiu@cs.hku.hk}, \\
  \texttt{zhenghaolin@stu.xmu.edu.cn}, 
  \texttt{chenlin@xmu.edu.cn},  \\
  \texttt{\{yegong, v-zhhang, jian.jiao,  nanduan, wzchen\}@microsoft.com} 
}

\begin{document}
\maketitle
\begin{abstract}
Many natural language processing (NLP) tasks rely on labeled data to train machine learning models with high performance. However, data annotation is time-consuming and expensive, especially when the task involves a large amount of data or requires specialized domains. 
Recently, GPT-3.5 series models have demonstrated remarkable few-shot and zero-shot ability across various NLP tasks. 
In this paper, we first claim that large language models (LLMs), such as GPT-3.5, can serve as an excellent crowdsourced annotator when provided with sufficient guidance and demonstrated examples. 
Accordingly, 
we propose AnnoLLM, 
an annotation system powered by LLMs, which adopts a two-step approach, \textit{explain-then-annotate}. 
Concretely, we first prompt LLMs to provide explanations for why the specific ground truth answer/label was assigned for a given example. 
Then, we construct the few-shot chain-of-thought prompt with the self-generated explanation and employ it to annotate the unlabeled data with LLMs. 
Our experiment results on three tasks, including user input and keyword relevance assessment, BoolQ, and WiC, demonstrate that AnnoLLM surpasses or performs on par with crowdsourced annotators. 
Furthermore, we build the first conversation-based information retrieval dataset employing AnnoLLM.  This dataset is designed to facilitate the development of retrieval models capable of retrieving pertinent documents for conversational text. Human evaluation has validated the dataset's high quality.

\end{abstract}

\section{Introduction}
Labeled data refers to a dataset that has been manually annotated with predefined target labels or categories. 
It is crucial to develop machine learning models for many NLP tasks, such as sentiment analysis \cite{socher-etal-2013-recursive}, machine translation \cite{seq2seq} and word sense disambiguation \cite{he-yiu-2022-controllable}. 
The process of labeling data is typically done by human annotators 
% who are given 
under 
specific guidelines and criteria on how to assign labels to each instance in the dataset. For example, in sentiment analysis, each sentence or document may be labeled with a polarity score such as ``positive'', ``negative'', or ``neutral''. 
However, it is very labor-intensive and time-consuming to create a large dataset with human annotation, 
which limits the availability of such data in various NLP tasks. 
% and the applicability of machine learning models in various NLP tasks. 

Previous works have shown that LLMs, such as GPT-3 \cite{NEURIPS2020_1457c0d6} and PaLM \cite{chowdhery2022palm}, 
achieve impressive results in many downstream tasks without requiring large-scale task-specific data or parameter tuning, but only with a few examples as instructions. 
OpenAI has recently launched the GPT-3.5 series models, the upgraded versions of GPT-3. 
% which are trained on a blend of text and code published before the end of 2021. 
Shortly after, OpenAI also unveiled ChatGPT, another fine-tuned version of GPT-3.5, which has gained significant global attention since its launch.  
% Within just two months since its release, ChatGPT has garnered a massive following of 100 million users worldwide, garnering significant global attention. 

Augmenting manually labeled data with pseudo-labeled data from GPT-3 is helpful for many NLP tasks, particularly when the labeling budget is restricted \cite{wang-etal-2021-want-reduce}. However, the quality of GPT-3's labeled data still lags behind that of manually labeled data. 
Considering the GPT-3.5 models' remarkable zero/few-shot capabilities, we raise an essential and significant inquiry: \textit{Can GPT-3.5 potentially replace crowdsourced annotators?}

Before answering this question, let us go over the process of crowdsourced data annotation. First, we need to provide  annotators with a specific definition of the task. Then, for classification tasks, we need to tell annotators the specific meanings of each category. Finally, we need to provide annotators with a few examples that have already been annotated as references. 
Naturally, we can guide GPT-3.5 to annotate data using the same approach as with human annotators by providing task definitions and example samples. 
Furthermore, we found that requesting LLMs to furnish the rationale behind the ground truth label for a particular example can prompt LLMs to produce high-quality explanations. 
Based on this, we create the few-shot chain-of-thought (COT) prompt \cite{wei2022chain} with the self-generated explanations to annotate data. 
We refer to this method as \textit{explain-then-annotate}, which further improves the annotation quality.

We summarize our contributions as follows: 
(1) We propose \textbf{AnnoLLM}, an \textbf{Anno}tation system powered by \textbf{L}arge \textbf{L}anguage \textbf{M}odels, which is based on \textit{explain-then-annotate} and has the potential to replace crowdsourced annotators to annotate data. 
% To enhance the data annotation capabilities of LLMs, we suggest a two-step approach called \textit{explain-then-annotate}. 
% In this approach, we leverage ChatGPT to generate a few-shot chain-of-thought prompt, which we then use to annotate unlabeled data. 
(2) Our results on three datasets verify the feasibility of substituting crowdsourced annotators with GPT-3.5, 
% \footnote{In this paper, we focus on using GPT-3.5 series models to annotate data for classification tasks.}
where it either surpasses or matches crowdsourced annotators. 
% (3) We construct the first conversation-based information retrieval dataset with our proposed AnnoLLM. Human evaluation shows that this dataset has high quality in terms of fluency, relevance, and factual consistency. 
(3) Furthermore, AnnoLLM is not limited to annotating classification data, and 
we create the first conversation-based information retrieval (\textbf{ConIR}) dataset using AnnoLLM\footnote{ConIR is available at: \url{https://github.com/NLPCode/AnnoLLM}.}. Through rigorous human evaluation, this dataset exhibits high quality in terms of fluency, relevance, and factual consistency.

\begin{figure*}
  \centering
    \includegraphics[width=1\textwidth]{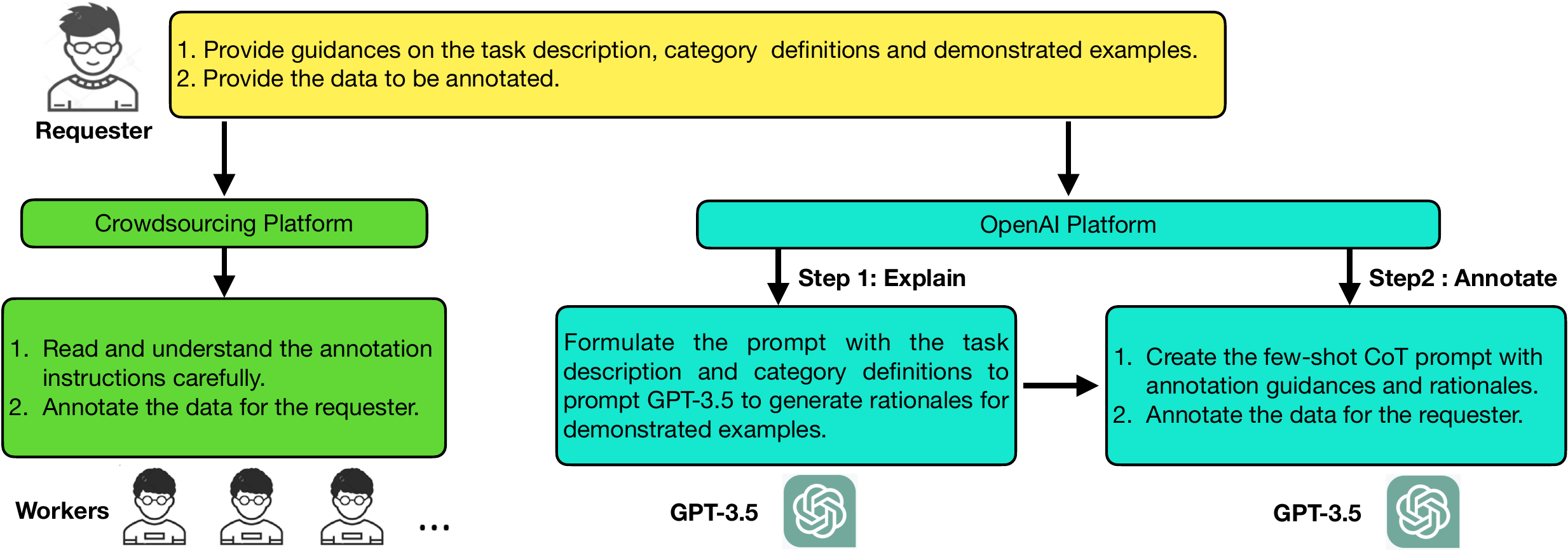} 
      \caption{ 
On the left is the annotation process used by crowdsourced workers, while on the right is AnnoLLM's process. AnnoLLM mimics the manual annotation process, with the exception that it generates explanations for each example before annotation. This ensures that each demonstrated example is accompanied by helpful explanations, making the annotation guidelines more informative and useful.
      }
      \label{fig.annotation}
\end{figure*}

\section{Approach}
Providing detailed instructions is crucial for crowdsourced workers to annotate data, as it helps them better understand task requirements and annotation standards, ultimately improving the quality and accuracy of annotated data. 
The instructions for each task mainly include three parts: task description, category definition, and demonstrated examples. 

Motivated by the guidance to human annotators, we will introduce how to convert GPT-3.5 into a zero-shot data annotator by providing guidance on the task description and category definitions in Section \ref{approach.zero}. Then, we will show how to transform GPT-3.5 into a few-shot data annotator using demonstrated examples in Section \ref{approach.few}. 
% For the ease of understanding, we show the crowdsourcing annotation and our proposed \textit{explain-then-annotate} processes in Figure \ref{fig.annotation}.
To make it easier to understand, we have provided a visual representation of the crowdsourcing annotation and AnnoLLM in Figure \ref{fig.annotation}.
Finally, in Section \ref{approach.create}, we will demonstrate the utilization of AnnoLLM for constructing the conversation-based information retrieval dataset.

\subsection{GPT-3.5 as a Zero-shot Data Annotator}\label{approach.zero}
In the zero-shot setting, we give the annotators only the task description and category definitions. 
The task description includes information on the task definition and purpose. Category definitions provide clear definitions for each category, so that the crowd workers can understand the meaning and standard of each category. 
Similarly, we provide GPT-3.5 with the task description and category definitions, allowing it to act as a zero-shot data annotator. 
We present the zero-shot prompts for GPT-3.5 on the user query and keyword relevance assessment (QK), WiC, and BoolQ tasks in Tables \ref{tab.qk.zero-shot}, \ref{tab.wic.zero-shot}, and \ref{tab.boolq.zero-shot}, respectively.

\subsection{GPT-3.5 as a Few-shot Data Annotator}\label{approach.few}
Providing labeled samples for each category can help annotators better understand how to annotate the data accurately. Similarly, we can also offer demonstrated examples to GPT-3.5, enabling it to serve as a few-shot annotator. 
We show the few-shot prompts for GPT-3.5 on QK, WiC, and BoolQ tasks in Tables \ref{tab.qk.few-shot}, \ref{tab.wic.few-shot}, and \ref{tab.boolq.few-shot}, respectively. 

Recent research \cite{wei2022chain} has discovered that adding human written rationales to demonstrated examples, called as chain-of-thought (CoT), can elicit LLMs' reasoning ability, thus gaining improvements on reasoning tasks. 
In this paper, we find that GPT-3.5\footnote{We resort to ChatGPT to generate explanations.} 
is proficient at generating reasonable explanations for demonstrated examples. 
% is a good reasoner who can automatically generate reasonable explanations for demonstrated examples. 
In the following, we will introduce how to generate explanations with GPT-3.5, and then create few-shot CoT prompts with the generated explanations.

\paragraph{Generating Explanations with GPT-3.5.} 
In this step, we simulate the human reasoning process to induce GPT-3.5 to explain the annotated examples. 
To be concrete, we present the task description, specific examplease, and the corresponding true labels to GPT-3.5, and then ask it to explain why the given label is appropriate for that example. By doing so, GPT-3.5 will generate reasonable explanations. 
For the QK task, we show how to use GPT-3.5 to explain why the label between the user query ``\textbf{google data studio sharepoint}'' and the keyword ``\textbf{sharepoint migration tool file share}'' is ``\textbf{Bad}'' in Table \ref{tab.qk.explanation} in Appendix \ref{sec.appendix.a}.  
Please refer to Table \ref{tab.wic.1} and Table \ref{tab.boolq.1} for how to generate explanations for the demonstrated examples of WiC and BoolQ.

\paragraph{Creating Few-shot CoT Prompts.}
We construct the few-shot CoT prompt using  the explanations generated by GPT-3.5. 
We show the few-shot CoT prompts on QK, WiC, and BoolQ tasks in Tables \ref{tab.qk.few-shot-cot}, \ref{tab.wic.few-shot-cot}, and \ref{tab.boolq.few-shot-cot} in Appendix \ref{sec.appendix.d}, respectively.

\subsection{GPT-3.5 as a Few-shot Data Creator}\label{approach.create}
AnnoLLM is not limited to labeling classification data. Next, we will introduce how we used AnnoLLM to construct the conversation-based information retrieval dataset. 
This dataset will facilitate the research and construction of conversation-based retrieval models.

Recently, ChatGPT, as a general artificial intelligence chatbot, has gained widespread attention, leading to the emergence of numerous information retrieval needs in the form of conversations. 
Specifically, during a conversation, users may ask questions that go beyond the knowledge scope of ChatGPT, requiring us to retrieve relevant literature from external knowledge bases. Traditional information retrieval datasets typically consist of queries $q$ and positive paragraphs $p$, denoted as $D = \{(q, p)\}$. 
We found that retrieval models trained on traditional datasets perform poorly on the conversation-based retrieval task (please refer to Section \ref{ConIR} for more details). 
This illustrates the necessity of constructing conversation-based retrieval datasets. 
Therefore, we propose to create a conversation-based information retrieval dataset. 

Conversation-based information retrieval aims to retrieve relevant passages from a large corpus for conversations. It is non-trivial to manually create datasets for this task. One intuitive idea is to use ChatGPT to generate a multi-turn conversation $c$ based on the query $q$ and the corresponding positive paragraph $p$, constructing a conversation dataset, $\{(c, p)\}$. 
However, we have found that this approach results in a dataset where a large portion of the conversation $c$ is directly copied from $p$. This is not desirable since it becomes easy to find $p$ related to $c$ based on word overlaps. 

To address this issue, we first utilize ChatGPT to enrich the given text paragraph $p$, obtaining $p^{\prime}$ (see Table \ref{tab.enrich_text.zero-shot}). Then, we generate the conversation $c$ based on the expanded paragraph $p^{\prime}$ and the given query $q$ (see Table \ref{tab.conversation_generation.zero-shot}). 
The expanded paragraph $p^{\prime}$ usually contains not only the information from the original paragraph $p$ but also some more detailed relevant information, while reducing the overlap of words with the original paragraph. 
In this way, the generated conversation $c$ can avoid having a large amount of identical text segments with the original paragraph $p$. However, since the expanded paragraph $p^{\prime}$ contains information beyond the original paragraph $p$, this may result in a relatively low relevance between the generated conversation $c$ and the original paragraph $p$. In other words, the original paragraph $p$ may not be a positive paragraph for the generated conversation $c$. 
Therefore, it is necessary to filter out the conversation instance $c$ that has low relevance to the original paragraph $p$. 
Due to the comparable data annotation capability of our proposed AnnoLLM, we naturally used AnnoLLM to judge whether the generated conversation $c$ and the original paragraph $p$ are related (see Table \ref{tab.data_filter.few-shot-cot}), and discarded data pairs that are irrelevant, resulting in the conversation-based information retrieval dataset.

\section{Experiment on Data Annotation}

\begin{table}
   % \footnotesize
  \scriptsize
  % \tiny
    \centering
      \begin{tabular}
        % {l|rrr|r}
      {
       m{0.14\textwidth}<{\raggedright}|
       m{0.07\textwidth}<{\raggedleft}
       m{0.07\textwidth}<{\raggedleft}
       m{0.07\textwidth}<{\raggedleft}
       }
      \toprule
      % \hline

      \textbf{Partition / Task}  & \textbf{QK} & \textbf{BoolQ} & \textbf{WiC}  \\
      \midrule
      % \hline
      Dev & 350 & 3270 & 638 \\
      Test & 1000 & 3245 & 1400 \\
    \bottomrule
    % \hline
    \end{tabular}
    \caption{Basic statistics of QK, BoolQ and WiC datasets. 
    }\label{tab.data}
\end{table}

% \begin{table}
%    \footnotesize
% %   \scriptsize
%   % \tiny
%     \centering
%       \begin{tabular}
%         % {l|rrr|r}
%       {
%        m{0.18\textwidth}<{\raggedright}|
%        m{0.1\textwidth}<{\raggedleft}
%        m{0.1\textwidth}<{\raggedleft}
%        }
%       % \toprule
%       \hline

%       Task / Partition & Dev & Test \\
%       % \midrule
%       \hline
%       QK & 350 &1000 \\
%       BoolQ& 3270 &3245\\   
%       WiC & 638 & 1400 \\ 

%     % \bottomrule
%     \hline
%     \end{tabular}
%     \caption{Basic statistics of QK, BoolQ and WiC datasets. 
%     }\label{tab.data}
% \end{table}
\subsection{Experimental Setups}
\paragraph{Datasets.} 
We evaluate AnnoLLM on three different tasks: QK, BoolQ, and WiC. The basic statistics of these datasets are shown in Table \ref{tab.data}. 
The \textbf{QK} task aims to judge whether the user input query is related to the given keywords. 
\textbf{BoolQ} (Boolean Questions) \cite{clark-etal-2019-boolq} is a question-answering task. In this task, each example comprises a brief passage and a yes/no question related to the passage. 
% The users of the Google search engine anonymously and without solicitation submit the questions, which are then matched with a paragraph from a Wikipedia article that provides the answer. 
The \textbf{WiC} (Word-in-Context) task \cite{pilehvar-camacho-collados-2019-wic} involves disambiguating word senses by classifying sentence pairs. 
% In this task, two text snippets are provided along with a polysemous word that occurs in both sentences. 
The goal is to determine if the target word shares the same sense in both sentences. 

\paragraph{Implementation Details.} 
We use ChatGPT (gpt-3.5-turbo) to generate explanations for demonstrated examples and implement AnnoLLM with text-davinci-003 (a powerful GPT-3.5 model). During generation, we set the temperature $t=0$ for text-davinci-003. 
% Since all tasks are binary classification, accuracy is used for evaluation.
As all tasks involve binary classification, accuracy is employed for evaluation.

\paragraph{Human Performances.}
To assess human performance on QK, we use UHRS\footnote{\url{https://prod.uhrs.playmsn.com/uhrs/}}, a crowdsourcing platform, for data annotation. 
% invite the crowdsourced annotators to annotate this data. 
Before annotation, we provide the task description, category definitions, and annotated examples to annotators. 
% Then, three annotators to label a data instance. 
If the annotated results of three workers are consistent, this result will be considered as the annotated label. 
Otherwise, additional annotators will continue annotating this data instance until three annotators have consistent annotation results. We require crowdsourced annotators to annotate all development and test sets. 
BoolQ and WiC are two of the most challenging datasets in superGLUE \cite{wang2019superglue}. 
For BoolQ, three authors labeled 110 randomly chosen examples, with human performance reaching 89\%.
As for WiC, \citet{pilehvar-camacho-collados-2019-wic} selected four groups of 100 test instances, and assigned each group to an annotator, achieving a human performance of 80\%.

\subsection{Experimental Results}
Table \ref{tab.qk.result} shows our experimental results on the QK development and test sets. Surprisingly, GPT-3.5 (text-davinci-003) performs worse in the few-shot setting compared to the zero-shot setting in this task. \citet{fu2022gptroadmap} speculate that the instruction tuning on GPT-3.5 may decrease its in-context learning ability but increase its zero-shot ability. On the other hand, AnnoLLM (text-davinci-003 + 4-shot CoT) outperforms its counterparts under the zero-shot and few-shot settings by around 6 and 8 points, respectively. Impressively, it even surpasses the crowdsourced annotators. 

Table \ref{tab.wic.result} presents our experimental results on WiC, from which we also see that AnnoLLM (text-davinci-003 + 8-shot CoT) outperforms its few-shot counterpart significantly. 
Nevertheless, there remains a considerable disparity between AnnoLLM and crowdsourced annotators. 
This can be attributed to the inherent complexity of the task, since even the best supervised models still exhibit a substantial gap compared to human performance.

As shown in Table \ref{tab.boolq.result}, AnnoLLM (text-davinci-003+8-shot CoT) surpasses human annotators and is comparable to supervised models on BoolQ, but does not show significant improvement compared to the few-shot method. However, this does not imply that CoT with generated explanation is not useful for this task. Section \ref{stability} shows that AnnoLLM with CoT exhibits better stability across different prompts, while its counterpart with the few-shot setting is highly sensitive to templates.

Overall, AnnoLLM surpasses or matches human performances in three tasks, demonstrating its potential to replace crowdsourced annotators. 
AnnoLLM differs from previous methods \cite{wei2022chain, wang2022self} in two aspects: (1) We use explanations generated by LLMs rather than those written by humans. 
(2) We have shown, for the first time, that the CoT method is effective in tasks beyond typical reasoning tasks.

\begin{table}[t] 
  \centering
 % \footnotesize
  \scriptsize
  % \tiny
   \begin{tabular}{
    m{0.3\textwidth}<{\raggedright}
    m{0.05\textwidth}<{\centering}
    m{0.05\textwidth}<{\centering}
    }
    \toprule
    % \hline
    \textbf{Models}   &  \textbf{Dev} & \textbf{Test} \\
    \midrule
    % \hline
    Crowdsourced Annotator    &   65.58  & 71.5 \\
    \midrule
    % \hline
    text-davinci-003 + zero-shot  & 67.71 & 70.00 \\
    text-davinci-003 + 8-shot  & 65.71 & 67.80 \\
    text-davinci-003 + 4-shot CoT (\textbf{AnnoLLM})   & \textbf{74.17}$^{\ast}$ & \textbf{75.60}$^{\ast}$ \\
    \bottomrule
    % \hline
 \end{tabular}
 \caption{Evaluation results (\%) on QK. 
 Accuracy is used as the evaluation metric. 
Results marked with $\ast$ represent the average result of five CoT prompts constructed with different generated explanations. 
}\label{tab.qk.result} 
\end{table}

\begin{table}[t] 
  \centering
 % \footnotesize
  \scriptsize
  % \tiny
   \begin{tabular}{
    m{0.3\textwidth}<{\raggedright}
    m{0.05\textwidth}<{\centering}
    m{0.05\textwidth}<{\centering}
    }
    \toprule
    % \hline
    \textbf{Models}   &  \textbf{Dev} & \textbf{Test} \\
    \midrule
    % \hline
    Crowdsourced Annotator    &   \multicolumn{2}{c}{80.0} \\
    \midrule
    % \hline
    \multicolumn{3}{l}{\textbf{Zero/Few-shot}} \\
    PaLM 540B + zero-shot &59.1$^{\ddagger}$ & - \\
    PaLM 540B + 5-shot  &64.6$^{\ddagger}$ & - \\
    text-davinci-003 + zero-shot  & 57.52 & 59.79 \\
    text-davinci-003 + 8-shot  & 67.71& 66.36 \\
    text-davinci-003 + 8-shot CoT (\textbf{AnnoLLM})  & \textbf{71.47}$^{\ast}$ & \textbf{69.17}$^{\ast}$ \\
    \midrule
    % \hline
    \multicolumn{3}{l}{\textbf{Fine-tune}} \\
    T5 11B \cite{raffel2020exploring}& 77.3$^\ddagger$ & 76.9$^{\dagger}$ \\
    PaLM 540B  & 78.8$^{\ddagger}$ & 77.4$^{\dagger}$ \\
    ST-MoE 32B \cite{zoph2022designing} & \textbf{81.0}$^{\ddagger}$   & \textbf{77.7}$^{\dagger}$\\
    \bottomrule
    % \hline
 \end{tabular}
 \caption{Evaluation results (\%) on the WiC task. 
 Accuracy is used as the evaluation metric. Results marked with $\dagger$ and $\ddagger$ are from the official SuperGLUE leaderboard\footnotemark[4] and  PaLM \cite{chowdhery2022palm}, respectively.  
 Results marked with $\ast$ represent the average result of five CoT prompts constructed with different generated explanations. 
 Numbers behind models denote the size of models'  parameters.
 }\label{tab.wic.result} 
\end{table} 

\footnotetext[4]{\url{https://super.gluebenchmark.com/leaderboard}}

\begin{table}[t] 
  \centering
 % \footnotesize
  \scriptsize
  % \tiny
   \begin{tabular}{
    m{0.32\textwidth}<{\raggedright}
    m{0.04\textwidth}<{\centering}
    m{0.04\textwidth}<{\centering}
    }
    \toprule
    % \hline
    \textbf{Models}   &  \textbf{Dev} & \textbf{Test} \\
    \midrule
    % \hline
    Crowdsourced Annotator    &  \multicolumn{2}{c}{89.0} \\
    \midrule
    % \hline
    \multicolumn{3}{l}{\textbf{Zero/Few-shot}} \\
    GPT-3 175B + zero-shot & 60.5 & -\\
    Gopher 280B + zero-shot \cite{rae2021scaling}& 79.3 & -\\
    Chinchilla 70B + zero-shot \cite{hoffmann2022training} &  83.7 & -\\
    PaLM 62B  + zero-shot &  84.8 & -\\
    PaLM 540B  + zero-shot &  88.0 & -\\
    LLaMA 65B  + zero-shot \cite{touvron2023llama}&  85.3  & -\\
    % \midrule
    text-davinci-003 + zero-shot & 84.28 & 84.30 \\
    text-davinci-003 + 8-shot  & 89.17& 89.10 \\
    text-davinci-003 + 8-shot CoT (\textbf{AnnoLLM})& \textbf{89.69} & \textbf{89.20} \\
    % \midrule
    % GPT-3.5 + zero-shot + inco & 85.44 & - \\
    % GPT-3.5 + 16-shot + inco  & 88.65& - \\
    % GPT-3.5 + 8-shot + CoT + inco  & 89.27 & -\\
    \midrule
    % \hline
    \multicolumn{3}{l}{\textbf{Fine-tune}} \\
    T5 11B \cite{raffel2020exploring}& 90.8$^\ddagger$ & 91.2$^{\dagger}$ \\
    PaLM 540B  & 92.2$^{\ddagger}$ & 91.9$^{\dagger}$ \\
    ST-MoE 32B \cite{zoph2022designing} & \textbf{93.1}$^{\ddagger}$   & \textbf{92.4}$^{\dagger}$\\
    
    \bottomrule
    % \hline
 \end{tabular}
 \caption{Evaluation results (\%) on the BoolQ task. Accuracy is used as the evaluation metric. Results marked with $\dagger$ and $\ddagger$ are from the official SuperGLUE leaderboard and PaLM, respectively. 
Numbers behind models denote the size of models' parameters.
 }\label{tab.boolq.result} 
\end{table}

\subsection{Ablation Study}
In this section, we conduct an experiment to compare the impact of various explanation generation methods on the performance of AnnoLLM.

Firstly, we want to investigate whether using ground truth labels is helpful for generating explanations for demonstrated examples. 
To answer this, we induce LLMs to generate explanations using prompts with and without ground truth labels. 
Specifically, we replace the last sentence of the prompt in Table \ref{tab.qk.explanation} \textit{Briefly explain why the relevance is "Bad"} with \textit{Briefly explain the relevance between the keyword and query} in Table \ref{tab.qk.explanation_wo_label}. 
From Table \ref{tab.ablation}, we found that not using true labels when generating explanations leads to a decrease in AnnoLLM's performance by approximately 3 points on the QK test set (row 4 vs. row 1).
This is because the model may generate explanations for incorrect answers without the guidance of ground truth labels. 
% Please refer to  and  for complete prompts and generated explanations.

% Based on the method in the fourth row, we filter out the incorrect explanations using true labels and retain three correct explanations for three out of four demonstrated samples. However, for the third sample, all five generated explanations are wrong, and we have to keep three incorrect explanations for this demonstrated example. That is why using the golden label to filter explanations does not bring significant improvement (row 5 vs. row 4). 

In Table \ref{tab.qk.explanation}, we found that LLMs initially reveal the true answer, and then provide an explanation for it. This differs from previous work \cite{wei2022chain}, where LLMs are prompted to give an explanation before outputting the answer. 
Therefore, we remove the initial sentence with labels from generated explanations (underlined text in Table \ref{tab.qk.few-shot-cot}). 
However, this modification does not lead to any improvement (row 2 vs. row 1). We speculate that this may be attributed to the disparity between our task and traditional reasoning tasks. 
In addition, we remove the last sentence containing the answer to the demonstrated examples (italicized text in Table \ref{tab.qk.few-shot-cot}), yet it does not have too much impact on the performance (row 3 vs. row 1). 
That is because the generated explanations already contain the correct answers. 
Nonetheless, to align with the format used in previous work \cite{wei2022chain}, we still append ground truth labels to generated explanations.

\begin{table}[t] 
  \centering
 % \footnotesize
  \scriptsize
  % \tiny
   \begin{tabular}{
   m{0.001\textwidth}<{\centering}|
    m{0.075\textwidth}<{\centering}|
    m{0.07\textwidth}<{\centering}|
    m{0.07\textwidth}<{\centering}|
    m{0.05\textwidth}<{\centering}
    m{0.05\textwidth}<{\centering}
    }
    % \hline
    \toprule
    \multicolumn{4}{c|}{\textbf{text-davinci-003 + 4-shot CoT}}  & \multicolumn{2}{c}{\textbf{Datasets}}\\
    \midrule
    \#&\textbf{Generate E with L} &\textbf{Delete L from E} & \textbf{Append L to E} & \textbf{Dev Set} & \textbf{Test Set} \\
    \midrule
    1&$\large\checkmark$ &   & $\large\checkmark$ & \textbf{74.17} & \textbf{75.60}  \\
    2&$\large\checkmark$ & $\large\checkmark$  & $\large\checkmark$ & 72.97 & 74.76 \\
    3&$\large\checkmark$ &  &  & 74.09 & 75.44  \\
    4&&   & $\large\checkmark$ & 72.63 & 72.84\\
    \bottomrule
 \end{tabular}
 \caption{Ablation study on the QK task. `E' and `L' refer to the generated explanations and ground truth labels, respectively. 
All results are averaged across five few-shot CoT prompts, each consisting of different generated explanations. 
% Results marked with a dagger ($\dagger$) represent the average outcome of three few-shot CoT prompts, each constructed with different generated explanations, while the remaining results represent the average of five.
}\label{tab.ablation} 
\end{table}

\subsection{More Analysis and Discussion}

\begin{figure*}
  \centering
    \subfigure[Consistency]{
      \centering
      \includegraphics[width=0.47\textwidth]{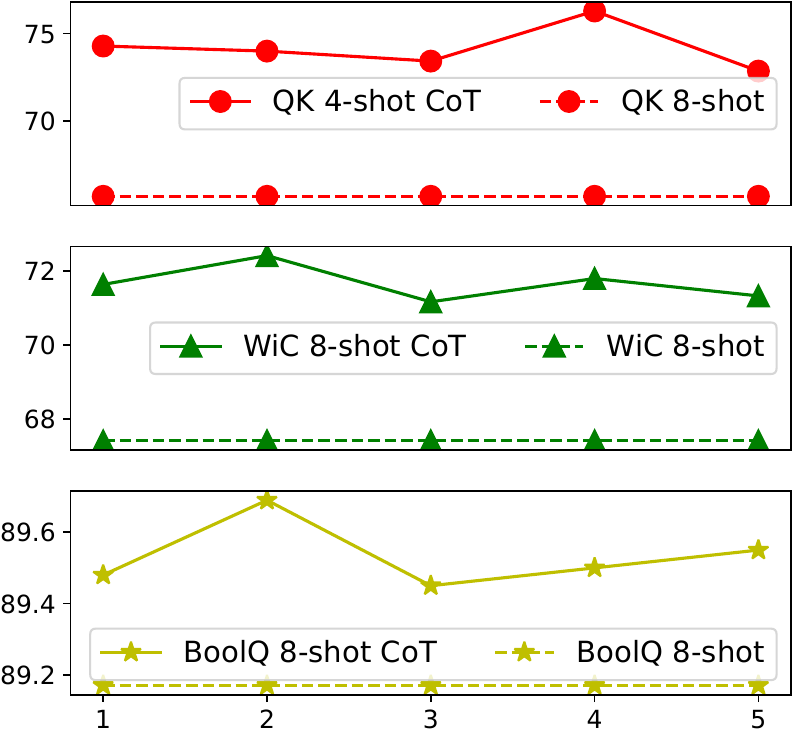} 
    }
    % \hspace{0mm}
      \subfigure[Stability]{
        \centering
        \includegraphics[width=0.284\textwidth]{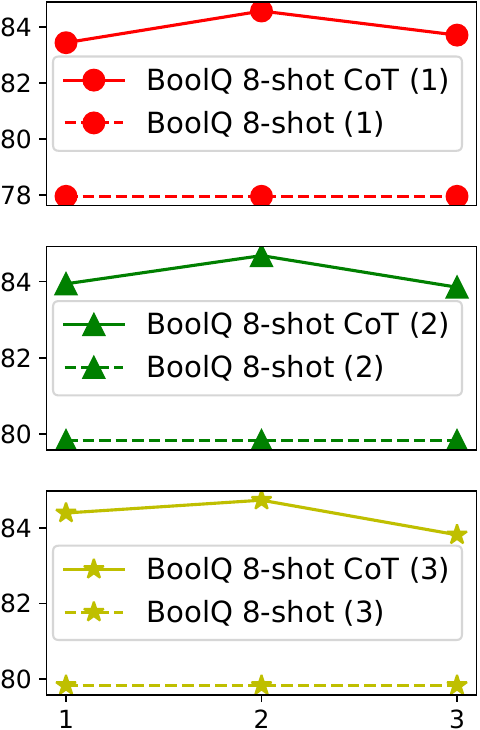} 
      }
      \caption{ 
        Subfigure (a) shows the performance on dev sets for CoT prompts created with different explanations. 
        Subfigure (b) shows the performance for different few-shot and few-shot CoT prompts on the dev set of BoolQ. 
        The X-axis represents the index of CoT prompts, while the Y-axis denotes accuracy.
        % X-axis: Index of CoT prompts. Y-axis: Accuracy.
      }
      \label{fig.consistency_stability}

\end{figure*}

\paragraph{Consistency Analysis of Generated Explanations.} \label{consistency}
In the ablation study, we found that the performance of AnnoLLM relies heavily on the generated explanations. 
This leads to a natural inquiry: \textit{Are the explanations produced by ChatGPT consistent enough for the same demonstrated sample?} 
To answer this, we generate five explanations for each sample, and obtain five different few-shot CoT prompts. 
As shown in Figure \ref{fig.consistency_stability} (a), these different few-shot CoT prompts yield similar performance in the QK, WiC, and BoolQ tasks. 
This indicates that the quality of the explanations generated by ChatGPT is sufficiently consistent.

% The results obtained using different few-shot CoT prompts are presented in Figure \ref{fig.consistency_stability} (a). It can be observed that different few-shot CoT prompts achieve similar performance in the QK, WiC, and BoolQ tasks. 

\paragraph{Stability Analysis of Generated Explanations.} \label{stability}
Figure \ref{fig.consistency_stability} (a) shows that AnnoLLM with few-shot CoT prompts significantly outperforms its counterpart with  few-shot settings  on QK and WiC. 
% with a notable margin of 5 percentage points on  QK and WiC in most cases. 
However, the improvement is quite modest on BoolQ, where it is generally less than 0.5. 
This does not mean that AnnoLLM with few-shot CoT prompts has no effect on BoolQ. 
To further analyze this, we make slight modifications to the existing prompts for BoolQ to obtain three few-shot CoT and few-shot prompts (refer to Appendix \ref{sec.appendix.e} for details). 
Figure \ref{fig.consistency_stability} (b) shows that the few-shot method is highly sensitive to templates. Even with slight modifications to templates, the experimental performances drop from around 89 to below 80 points. 
In comparison, AnnoLLM with few-shot CoT prompts suffers less performance loss, which  outperforms its counterpart with few-shot templates by around 4 points. 
To summarize, the few-shot setting is more picky about templates, whereas few-shot CoT exhibits better stability across different templates.

\section{Experiment on Data Creation}\label{ConIR}
% \subsection{Experimental Setups}
\paragraph{Datasets.} 
We construct the conversation-based information retrieval (\textbf{ConIR}) dataset based on the MS-MARCO passage ranking dataset \cite{bajaj2016ms}. 
The sizes of the training and test sets for ConIR are 71,557 and 3,000 respectively. 
\paragraph{Implementation Details.} Since ChatGPT is optimized for chat, we use it to create ConIR, namely using it to enrich paragraphs, generate and filter out irrelevant conversations in Appendix \ref{sec.appendix.f}. 
Following previous work \cite{qu-etal-2021-rocketqa}, we resort to MRR@10 and Recall of top-k (R@k) to evaluate the retrieval performance on different models. 
% which represents the proportion of top k retrieved passages that contain the answers.

% \subsection{Experimental Results} 
\paragraph{Zero-shot Performance.} We train two typical dense retrieval models, DPR \cite{karpukhin-etal-2020-dense} (initialized with DistilBERT \cite{DistilBERT}) and PROD \cite{lin2023prod}, on MS-MARCO, and then evaluate them on the test set of ConIR. Notably, both models exhibit poor performance on ConIR, as demonstrated in Table \ref{table.ConIR}. 
This indicates that dense retrieval models trained on traditional datasets are not directly applicable to conversation-based information retrieval.

\paragraph{In-domain Performance. } As shown in Table \ref{table.ConIR}, DPR fine-tuned on the training set of ConIR performs much better than its zero-shot counterpart, highlighting the necessity of the ConIR dataset. 

\paragraph{Human Evaluation. } We randomly select 100 generated conversations and their paired paragraphs. Three annotators are asked to assess the fluency of conversations on a 5-point Likert scale from 1 (not fluent) to 5 (extremely fluent), and their relevance and factual consistency with the paired passages on a 3-point Likert scale. 
Table \ref{table.ConIR.human} shows that the  conversations of ConIR 
exhibit remarkable fluency, displaying a strong correlation with the paired paragraphs in terms of relevance and factual consistency. 
% are very fluent, and have a strong relevance and consistency with the paired paragraphs. 
The inter-annotator agreement measured using Fleiss' \textit{kappa} \cite{Fleiss1971MeasuringNS} is 0.55, implying moderate agreement \cite{Landis1977TheMO}.
Please refer to Appendix \ref{sec.appendix.g} for more details.
% about human evaluation.

\begin{table}
\scriptsize
  \centering
    \begin{tabular}{
     m{0.12\textwidth}<{\raggedright}
     m{0.05\textwidth}<{\centering}
     m{0.04\textwidth}<{\centering}
     m{0.035\textwidth}<{\centering}
     m{0.035\textwidth}<{\centering}
     m{0.04\textwidth}<{\centering}
     }
    \toprule
    \textbf{Models}& \textbf{MRR@10} &\textbf{R@1}  &\textbf{R@5} &\textbf{R@50} &\textbf{R@100} \\
    \midrule 
    % \multicolumn{3}{l}{\textbf{Zero-shot}} \\
    \textbf{DPR (Zero-shot)} & 7.01 &4.85 & 9.75& 18.70 & 22.08\\
    \textbf{PROD (Zero-shot)} & 10.61 & 7.53 & 14.80 & 28.22 & 32.77 \\
    \midrule
    % \multicolumn{3}{l}{\textbf{Fine-tune}} \\
    \textbf{DPR (Fine-tune)} & \textbf{19.32} & \textbf{12.27} & \textbf{28.60}  & \textbf{56.13} &  \textbf{64.25}\\
    \bottomrule

  \end{tabular}
  \caption{Retrieval results on the test set of ConIR. 
  }\label{table.ConIR}
\end{table}

\begin{table}
\scriptsize
  \centering
    \begin{tabular}{
     m{0.05\textwidth}<{\raggedright}
     m{0.05\textwidth}<{\centering}
     m{0.05\textwidth}<{\centering}
     m{0.2\textwidth}<{\centering}
     }
    \toprule
    \textbf{Fluency}  &\textbf{Relevance} &\textbf{Consistency} & \textbf{Inter-annotator agreement} \\
    \midrule 
    4.99 & 2.53& 2.41& 0.55\\
    \bottomrule

  \end{tabular}
  \caption{Human evaluation results on ConIR.
  }\label{table.ConIR.human}
\end{table}
\section{Related Work}
\paragraph{Large-scale Language Models.}
GPT (Generative Pre-trained Transformer) is a family of language models developed by OpenAI, designed to generate human-like natural language text. 
GPT models are based on the Transformer architecture \cite{NIPS2017_3f5ee243}, which are pre-trained on an enormous corpus of text by predicting the next token based on the previous context. 
Over the years, OpenAI has continuously increased the parameters and training data of its models, and has released GPT \cite{Radford2018ImprovingLU}, GPT-2 \cite{Radford2019LanguageMA}, and GPT-3 \cite{NEURIPS2020_1457c0d6} from 2018 to 2020.
% GPT-3 is a powerful language model with 175 billion parameters, and make a significant advance in the field of NLP. 
One unique feature of GPT-3 is in-context learning, where one can apply it to various tasks by simply providing few-shot demonstrations without any fine-tuning. 
Furthermore, OpenAI fine-tuned GPT-3 on the code data or instruction data, releasing Codex \cite{chen2021evaluating} and InstructGPT \cite{ouyangtraining}, respectively. 
Recently, OpenAI released GPT-3.5 series models, including text-davinci-003 and ChatGPT, by training on text and code data, then tuning with supervised instructions and reinforcement learning with human feedback.  Recent research has shown that GPT-3.5 has strong few-shot and zero-shot learning abilities on various NLP tasks \cite{jiao2023chatgpt, wei2023zero}.

In this paper, we first propose that we can readily change GPT-3.5 to a good data annotator for a specific task by providing the detailed annotation instructions similar to human annotators. 

\paragraph{Pseudo Annotated Data.} 
% Creating human-annotated data is tedious and costly, particularly for complex tasks or specialized  domains where there may not be sufficient data available. 

Creating pseudo-annotated data is commonly used to generate labeled data for a specific task when there is a limited amount of annotated data available.
Back-translation involves translating a target language sentence back into the source language, which is first proposed to improve neural machine translation models with synthetic parallel data \cite{sennrich-etal-2016-improving}. 
Beyond machine translation, this technique has also been applied to unsupervised text style transfer \cite{prabhumoye-etal-2018-style} and image style transfer \cite{zhu2017unpaired}.
In addition, rule-based methods are widely used to construct synthetic data. 
For example, \citet{zhang2020pegasus} resorted to the lead bias to create paired data to pre-train the text summarization model, PEGASUS. 
 \citet{lee-etal-2019-latent} pre-trained the retriever with the Inverse Cloze Task, which aims to predict the context based on the given sentence. 
However, these methods are task-specific and difficult to generalize to other tasks. 
% In this paper, we study how to transform the powerful language model GPT-3.5 into a general-purpose data annotator. 
This paper explores the transformation of GPT-3.5 into a versatile data annotator. 
By providing the corresponding task description and few-shot CoT demonstrations, GPT-3.5 can easily annotate data for various tasks. Inspired by AnnoLLM, \citet{he-etal-2023-pivotfec} employed LLMs to introduce factual errors into accurate text, thereby generating data for factual error correction \cite{thorne-vlachos-2021-evidence, he2024improving}.

\section{Conclusion}
In this paper, we present AnnoLLM, a novel annotation system powered by LLMs that has the potential to replace traditional crowdsourced annotators. 
AnnoLLM adopts a two-step approach, \textit{explain-then-annotate}. In this method, LLMs are initially employed to generate a few-shot CoT prompt, which is subsequently utilized to prompt LLMs in annotating unlabeled data. 
Our experimental results on three datasets demonstrate the feasibility of using AnnoLLM to substitute crowdsourced annotators. 
Moreover, we introduce the ConIR dataset, which is created using AnnoLLM, to facilitate the research on  conversation-based information retrieval. 

% Furthermore, we create the high-quality ConIR dataset for conversation-based information retrieval with AnnoLLM. 
% which highlights the potential to facilitate the development of using LLMs like GPT-3.5 to annotate data for various NLP tasks. 

\section{Acknowledgments}
This work is supported by HKU-SCF FinTech Academy, Shenzhen-Hong Kong-Macao Science and Technology Plan Project (Category C Project: SGDX20210823103537030), and Theme-based Research Scheme of RGC, Hong Kong (T35-710/20-R). 
We would like to thank the anonymous reviewers for their constructive and informative feedback on this work.
\bibliography{anthology,custom}
\bibliographystyle{acl_natbib}

\clearpage

\appendix

\section{Prompts for Generating Explanations}\label{sec.appendix.a}
\begin{table*}[b] 
  \centering
 \footnotesize
%   \scriptsize
  % \tiny
   \begin{tabular}{
    m{0.1\textwidth}<{\centering}|
    m{0.84\textwidth}
    }
    \toprule
    \textbf{Input}  & Directions: Given a search engine query: "\textbf{google data studio sharepoint}", first, consider what the user could have in mind when they type in the query and allow for misspellings or other ambiguity, then classify the relevance of keyword: "\textbf{sharepoint migration tool file share}" to the query into one of the following categories: "Not bad", "Bad". The definitions of the categories are\newline
"Not bad": the keyword is relevant to the user's search query. This can include: broader or narrower product selection, competitor or alternative products, accessories, products often purchased together and related topics as well as direct matches to the user's search.\newline
"Bad": the keyword is not relevant to the user's search query. There is no relationship between the query and keyword. This includes but is not limited to: incorrect/unrelated product, unrelated topic, wrong location when location is important, cannot be used in place of query product nor are they commonly purchased together, etc.\newline
Briefly explain why the relevance is "\textbf{Bad}", with a response length not exceeding 100 words.   \\
    \midrule
    \textbf{Output 1}  & The relevance of the keyword "sharepoint migration tool file share" to the search query "google data studio sharepoint" is "Bad". This keyword is not relevant to the user's search query as it pertains to a different tool or technology (SharePoint) that is not mentioned in the search query. The keyword focuses on a migration tool for SharePoint file sharing, while the search query is focused on Google Data Studio. There is no relationship between the two, so the keyword is not relevant to the user's search query.\\
    \midrule
    \textbf{Output 2}  & The relevance of "sharepoint migration tool file share" to the search query "google data studio sharepoint" is "Bad." The keyword is not relevant to the user's search query because it deals with SharePoint migration and file sharing, whereas the user's query is focused on Google Data Studio and SharePoint. There is no relationship between these two topics, and the keyword cannot be used in place of the query product.\\
    \midrule
    \textbf{Output 3}  & The relevance is "Bad". The keyword "sharepoint migration tool file share" is not directly related to the query "google data studio sharepoint". The keyword refers to tools and services for migrating data from SharePoint to another file share platform, whereas the query is about using Google Data Studio to connect with SharePoint data. There is no clear relationship between the two topics.\\
    \midrule
    \textbf{Output 4}  & The relevance of the keyword "sharepoint migration tool file share" to the search query "google data studio sharepoint" is "Bad". The keyword is about a tool for migrating data from SharePoint, which is not directly related to Google Data Studio. Google Data Studio is a data visualization and reporting tool, not a migration tool for SharePoint.\\
    \midrule
    \textbf{Output 5}  & The relevance is "Bad" as the keyword "sharepoint migration tool file share" has no direct relation to the user's search query "google data studio sharepoint". The keyword pertains to a tool used for migrating data within a SharePoint platform, while the query is focused on Google Data Studio and SharePoint. These two concepts are not directly related and therefore, the keyword is not relevant to the user's search query.\\
    \bottomrule
 \end{tabular}
 \caption
 { Illustration of how to use a LLM to generate explanations for the QK task. 
``Input'' and ``Output'' refer to the prompt provided to the LLM and generated explanations, respectively.}\label{tab.qk.explanation} 
\end{table*}

\begin{table*}[hbp!]
  \centering
 \footnotesize
  % \scriptsize
  % \tiny
   \begin{tabular}{
    m{0.1\textwidth}<{\centering}|
    m{0.8\textwidth}
    }
    \toprule
    \textbf{Input}  & Task description: The task is to identify if the occurrences of "place" in `\textbf{Do you want to come over to my "place" later?}' and `\textbf{A political system with no "place" for the less prominent groups.}' correspond to the same meaning or not. The target word "place" in the first sentence and the second sentence is enclosed in quotation marks. There is no need to consider the tense and form of "place". If "place" in the first sentence and the second sentence correspond to the same meaning, output "true", otherwise output "false".\newline
Briefly explain why the answer is "\textbf{false}", with a response length not exceeding 100 words.\\
    \midrule
    \textbf{Output 1}  & The answer is "false". In the first sentence, "place" refers to a physical location or a person's home. In the second sentence, "place" refers to a position or status in a political system. Therefore, although the spelling and punctuation of "place" are the same in both sentences, the context and meaning of the word are different.\\
    \midrule
    \textbf{Output 2}  & The answer is "false". In the first sentence, "place" refers to a physical location, meaning a house or an apartment. In contrast, in the second sentence, "place" refers to a figurative location, meaning a position or role within a political system. Therefore, the two occurrences of "place" have different meanings and do not correspond to the same meaning.\\
    \midrule
    \textbf{Output 3}  & The answer is "false". In the first sentence, "place" refers to a physical location or someone's home, while in the second sentence, "place" refers to a position or status in a political system. These are different meanings, and therefore, the occurrences of "place" in the two sentences do not correspond to the same meaning.\\
    \midrule
    \textbf{Output 4}  &The answer is "false". In the first sentence, "place" refers to a physical location, as in a person's home. In contrast, in the second sentence, "place" refers to a position or status within a system or society, as in a group's role or standing. Thus, the two uses of "place" have different meanings, and do not correspond to the same meaning.\\
    \midrule
    \textbf{Output 5}  & The answer is "false". In the first sentence, "place" refers to a physical location where the speaker is inviting someone to come over to. In contrast, in the second sentence, "place" refers to a position or role in a political system. Therefore, the two occurrences of "place" have different meanings and cannot be considered to correspond to the same meaning.
\\
    \bottomrule
 \end{tabular}
 \caption{Illustration of how to use LLM to generate explanations for the WiC task. 
``Input'' and ``Output'' refer to the prompt provided to the LLM and generated explanations, respectively.}\label{tab.wic.1} 
\end{table*}

\begin{table*}
  \centering
 \footnotesize
  % \scriptsize
  % \tiny
   \begin{tabular}{
    m{0.1\textwidth}<{\centering}|
    m{0.84\textwidth}
    }
    \toprule
    \textbf{Input}  & Task description: Based on the information provided in the passage: "\textbf{The Elder Scrolls Online -- As with other games in The Elder Scrolls series, the game is set on the continent of Tamriel. The events of the game occur a millennium before those of The Elder Scrolls V: Skyrim and around 800 years before The Elder Scrolls III: Morrowind and The Elder Scrolls IV: Oblivion. It has a broadly similar structure to Skyrim, with two separate conflicts progressing at the same time, one with the fate of the world in the balance, and one where the prize is supreme power on Tamriel. In The Elder Scrolls Online, the first struggle is against the Daedric Prince Molag Bal, who is attempting to meld the plane of Mundus with his realm of Coldharbour, and the second is to capture the vacant imperial throne, contested by three alliances of the mortal races. The player character has been sacrificed to Molag Bal, and Molag Bal has stolen their soul, the recovery of which is the primary game objective.}", determine if the answer to the question: "\textbf{is elder scrolls online the same as skyrim}" is "yes" or "no". If the answer is affirmative, indicate "true", otherwise indicate "false".\newline
Briefly explain why the answer is "\textbf{false}", with a response length not exceeding 100 words.   \\
    \midrule
    \textbf{Output 1}  & The answer is "false". Although The Elder Scrolls Online is set on the continent of Tamriel, like Skyrim and other games in The Elder Scrolls series, it takes place a millennium before Skyrim and around 800 years before Morrowind and Oblivion. While it has a similar structure to Skyrim, with two conflicts progressing at the same time, the primary struggle in The Elder Scrolls Online is against the Daedric Prince Molag Bal, who has stolen the player character's soul, and the second is to capture the vacant imperial throne. Therefore, The Elder Scrolls Online is not the same game as Skyrim.\\
    \midrule
    \textbf{Output 2}  & The answer is "false". While The Elder Scrolls Online is set in the same continent of Tamriel as Skyrim, and shares a similar structure with two conflicts happening simultaneously, the events of the game occur a millennium before Skyrim, and its primary antagonist is the Daedric Prince Molag Bal, not Alduin as in Skyrim. The gameplay mechanics and overall experience of The Elder Scrolls Online are also distinct from Skyrim, as it is a massively multiplayer online role-playing game (MMORPG), while Skyrim is a single-player game.\\
    \midrule
    \textbf{Output 3}  & The answer is "false". While The Elder Scrolls Online and Skyrim share the same fictional universe and are part of The Elder Scrolls series, they are not the same game. The Elder Scrolls Online is set a millennium before Skyrim and has a different storyline and gameplay mechanics. In The Elder Scrolls Online, the player fights against Molag Bal and attempts to recover their stolen soul, while in Skyrim, the player character is a Dragonborn who fights against Alduin, the World-Eater. Additionally, The Elder Scrolls Online is a massively multiplayer online game, while Skyrim is a single-player game.\\
    \midrule
    \textbf{Output 4}  &The answer is "false". While The Elder Scrolls Online and Skyrim share some similarities, such as being set in the same continent of Tamriel and having a similar structure, they are different games with different stories and timelines. The events in The Elder Scrolls Online take place a millennium before Skyrim and revolve around two conflicts, one against the Daedric Prince Molag Bal and the other for the vacant imperial throne, while Skyrim has its own unique story centered around the return of the dragons to the world.\\
    \midrule
    \textbf{Output 5}  & The answer is "false". While The Elder Scrolls Online and Skyrim share the same fictional world of Tamriel, they are not the same game. The events of The Elder Scrolls Online take place a millennium before Skyrim, and the game features a different storyline and game structure. In The Elder Scrolls Online, the player character is sacrificed to the Daedric Prince Molag Bal and must recover their soul, while in Skyrim, the player character is a Dragonborn on a quest to defeat the dragon Alduin and save Tamriel from destruction.\\
    \bottomrule
 \end{tabular}
 \caption{Illustration of how to use a LLM to generate explanations for the BoolQ task. 
``Input'' and ``Output'' refer to the prompt provided to the LLM and generated explanations, respectively.
}\label{tab.boolq.1} 
\end{table*}

\begin{table*}
  \centering
 \footnotesize
  % \scriptsize
  % \tiny
   \begin{tabular}{
    m{0.1\textwidth}<{\centering}|
    m{0.84\textwidth}
    }
    \toprule
    \textbf{Input}  & Directions: Given a search engine query: "\textbf{google data studio sharepoint}", first, consider what the user could have in mind when they type in the query and allow for misspellings or other ambiguity, then classify the relevance of keyword: "\textbf{sharepoint migration tool file share}" to the query into one of the following categories: "Not bad", "Bad". The definitions of the categories are\newline
"Not bad": the keyword is relevant to the user's search query. This can include: broader or narrower product selection, competitor or alternative products, accessories, products often purchased together and related topics as well as direct matches to the user's search.\newline
"Bad": the keyword is not relevant to the user's search query. There is no relationship between the query and keyword. This includes but is not limited to: incorrect/unrelated product, unrelated topic, wrong location when location is important, cannot be used in place of query product nor are they commonly purchased together, etc.\newline
Briefly explain the relevance between the keyword and query, with a response length not exceeding 100 words.  \\
    \midrule
    \textbf{Output 1}  & The user's query "google data studio sharepoint" suggests they are looking for information on how to integrate or connect Google Data Studio with SharePoint, possibly for data visualization or reporting purposes. The keyword "sharepoint migration tool file share" is not directly relevant to the user's query as it pertains to a tool used for migrating files between SharePoint instances, rather than integrating Google Data Studio with SharePoint. Therefore, the keyword is considered "Bad" in relation to the user's query.\\
    \midrule
    \textbf{Output 2}  & When a user searches for "Google Data Studio SharePoint," they are likely looking for information on how to integrate Google Data Studio with SharePoint or how to use Google Data Studio to create reports on SharePoint data. The keyword "SharePoint migration tool file share" is not directly relevant to the query as it pertains to a tool for migrating files to SharePoint rather than using Google Data Studio with SharePoint. Therefore, the keyword is considered "bad" in this context.\\
    \midrule
    \textbf{Output 3}  & \textcolor{red}{When a user searches for "google data studio sharepoint," they are likely looking for information on how to integrate or connect Google Data Studio with SharePoint, a popular document management and collaboration platform. The keyword "sharepoint migration tool file share" is somewhat relevant to the query, as it pertains to migrating files from one SharePoint instance to another, which could be useful information for someone looking to integrate the two platforms. Therefore, I would classify this keyword as "Not bad."}\\
    \midrule
    \textbf{Output 4}  & The search query "google data studio sharepoint" likely suggests that the user is looking for information about how to integrate or use Google Data Studio with SharePoint. They may be interested in creating reports or visualizations using data from SharePoint in Google Data Studio. The keyword "sharepoint migration tool file share" is not relevant to the query as it refers to a tool for migrating files from one SharePoint site to another, and does not address the query's focus on integrating Google Data Studio with SharePoint. Therefore, the keyword is considered "Bad" for this search query.\\
    \midrule
    \textbf{Output 5}  & The search engine query "google data studio sharepoint" suggests that the user may be looking for information on how to integrate or use Google Data Studio with SharePoint, which is a web-based collaborative platform used for document management and storage. The keyword "sharepoint migration tool file share" is not directly relevant to the user's search query, as it is related to a tool used for migrating files between different file-sharing platforms, and does not address the user's original intent of using Google Data Studio with SharePoint. Therefore, the keyword is classified as "Bad."\\
    \bottomrule
 \end{tabular}
 \caption{Illustration of how to use a LLM to generate explanations for the user query and keyword relevance assessment task without using the ground truth labels. 
``Input'' and ``Output'' refer to the prompt provided to the LLM and the generated explanations, respectively.
 The red outputs indicate incorrect generated explanations.
 }\label{tab.qk.explanation_wo_label} 
\end{table*}

\clearpage
\section{Zero-shot Prompts}\label{sec.appendix.b}
\begin{table*}[hbp!] 
  \centering
 \footnotesize
  % \scriptsize
  % \tiny
   \begin{tabular}{
    m{0.96\textwidth}
    }
    \toprule
    Given a search engine query, first, consider what the user could have in mind when they type in the query and allow for misspellings or other ambiguity, then classify the relevance of keyword to the query into one of the following categories: "Not bad", or "Bad". The definitions of the categories are\\
    "Not bad": the keyword is relevant to the user's search query. This can include: broader or narrower product selection, competitor or alternative products, accessories, products often purchased together and related topics as well as direct matches to the user's search.\\
    "Bad": the keyword is not relevant to the user's search query. There is no relationship between the query and keyword. This includes but is not limited to: incorrect/unrelated product, unrelated topic, wrong location when location is important, cannot be used in place of query product nor are they commonly purchased together, etc.\\
    Please predict whether the keyword is relevant to the query or not. The answer should be exact "Not bad" or "Bad".\\
    \\
    Query: \textbf{\{query\}}\\
    Keyword: \textbf{\{keyword\}}\\
    Answer: \\
    \bottomrule
 \end{tabular}
 \caption{Zero-shot prompt for the QK task.}\label{tab.qk.zero-shot} 
\end{table*} 

\begin{table*}[hbp!]
  \centering
 \footnotesize
  % \scriptsize
  % \tiny
   \begin{tabular}{
    m{0.96\textwidth}
    }
    \toprule
    The goal of this task is to determine whether the targeted word in the first sentence and the second sentence conveys the same meaning. Please note that if the targeted word appears multiple times in the sentences, only the instance of the word surrounded by quotation marks should be considered. Additionally, the tense and form of the targeted word should not be taken into account. If the targeted word in the first sentence and the second sentence corresponds to the same meaning, output "True"; otherwise, output "False".\\
    To complete this task, you will need to predict whether the targeted word "w" in the first sentence "s1" and the second sentence "s2" convey the same meaning. Your answer should be either "True" or "False".\\
    \\
    w: \textbf{\{target word\}}\\
    s1: \textbf{\{first sentence\}}\\
    s2: \textbf{\{second sentence\}}\\
    Answer: \\
    \bottomrule
 \end{tabular}
 \caption{Zero-shot prompt for the WiC task.}\label{tab.wic.zero-shot} 
\end{table*} 

\begin{table*}[hbp!]
  \centering
 \footnotesize
  % \scriptsize
  % \tiny
   \begin{tabular}{
    m{0.96\textwidth}
    }
    \toprule
    Yes/No question-answering consists of a short passage and a Yes/No question about the passage. The questions are provided anonymously and unsolicited by users of the Google search engine, and afterwards paired with a paragraph from a Wikipedia article containing the answer. If there exists evidence in the passage that supports the facts in the question, the answer should be "Yes". If there exists evidence in the passage that denies the facts in the question, the answer should be "No".\\
    Your task is to read the passage and predict whether the answer to the question is "Yes" or "No".\\
    \\
    Passage: \textbf{\{passage\}}\\
    Question: \textbf{\{question\}}\\
    Answer: \\
    \bottomrule
 \end{tabular}
 \caption{Zero-shot prompt for the BoolQ task.}\label{tab.boolq.zero-shot} 
\end{table*} 

\clearpage
\section{Few-shot Prompts}\label{sec.appendix.c}

\begin{table}[!hbp] 
  \centering
 \footnotesize
  % \scriptsize
  % \tiny
   \begin{tabular}{
    m{0.46\textwidth}
    }
    \toprule
    Given a search engine query, first, consider what the user could have in mind when they type in the query and allow for misspellings or other ambiguity, then classify the relevance of keyword to the query into one of the following categories: "Not bad", or "Bad". The definitions of the categories are\\
    "Not bad": the keyword is relevant to the user's search query. This can include: broader or narrower product selection, competitor or alternative products, accessories, products often purchased together and related topics as well as direct matches to the user's search.\\
    "Bad": the keyword is not relevant to the user's search query. There is no relationship between the query and keyword. This includes but is not limited to: incorrect/unrelated product, unrelated topic, wrong location when location is important, cannot be used in place of query product nor are they commonly purchased together, etc.\\
    Please predict whether the keyword is relevant to the query or not. The answer should be exact "Not bad" or "Bad".\\
    \\
    Query: google data studio sharepoint\\
    Keyword: sharepoint migration tool file share\\
    Answer: Bad\\
    \\
    Query: motorhomes sale\\
    Keyword: rv sale used class c\\
    Answer: Not bad\\
    \\
    Query: southern exposure seed exchange company\\
    Keyword: uk poppy seeds\\
    Answer: Not bad\\
    \\
    Query: nissan parts canada\\
    Keyword: purchase tires\\
    Answer: Bad\\
    \\
    Query: alcohol detoxing\\
    Keyword: inpatient drug rehab\\
    Answer: Not bad\\
    \\
    Query: loudmouth clothing sale\\
    Keyword: levis jeans\\
    Answer: Bad\\
    \\
    Query: firefox mac sierra\\
    Keyword: opera browser mac\\
    Answer: Not bad\\
    \\
    Query: google images\\
    Keyword: buy photo\\
    Answer: Bad\\
    \\
    Query: \textbf{\{query\}}\\
    Keyword: \textbf{\{keyword\}}\\
    Answer: \\
    \bottomrule
 \end{tabular}
 \caption{Few-shot exemplars prompt for the QK task.}\label{tab.qk.few-shot} 
\end{table}

\begin{table}[!hbp] 
  \centering
 \footnotesize
  % \scriptsize
  % \tiny
   \begin{tabular}{
    m{0.46\textwidth}
    }
    \toprule
    The goal of this task is to determine whether the targeted word in the first sentence and the second sentence conveys the same meaning. Please note that if the targeted word appears multiple times in the sentences, only the instance of the word surrounded by quotation marks should be considered. Additionally, the tense and form of the targeted word should not be taken into account. If the targeted word in the first sentence and the second sentence corresponds to the same meaning, output "True"; otherwise, output "False".\\
    To complete this task, you will need to predict whether the targeted word "w" in the first sentence "s1" and the second sentence "s2" convey the same meaning. Your answer should be either "True" or "False".\\
    \\
    w: "place"\\
    s1: Do you want to come over to my "place" later?\\
    s2: A political system with no "place" for the less prominent groups.\\
    Answer: False\\
    \\
    w: "hold"\\
    s1: The general ordered the colonel to "hold" his position at all costs.\\
    s2: "Hold" the taxi.\\
    Answer: True\\
    \\
    w: "summer"\\
    s1: We like to "summer" in the Mediterranean.\\
    s2: We "summered" in Kashmir.\\
    Answer: True\\
    \\
    w: "approach"\\
    s1: "Approach" a task.\\
    s2: To "approach" the city.\\
    Answer: False\\
    \\
    w: "run"\\
    s1: "Run" rogue.\\
    s2: She "ran" 10 miles that day.\\
    Answer: False\\
    \\
    w: "head"\\
    s1: His horse won by a "head".\\
    s2: He is two "heads" taller than his little sister.\\
    Answer: True\\
    \\
    w: "meet"\\
    s1: The company agrees to "meet" the cost of any repairs.\\
    s2: This proposal "meets" my requirements.\\
    Answer: True\\
    \\
    w: "development"\\
    s1: The organism has reached a crucial stage in its "development".\\
    s2: Our news team brings you the latest "developments".\\
    Answer: False\\
    \\
    w: \textbf{\{target word\}}\\
    s1: \textbf{\{first sentence\}}\\
    s2: \textbf{\{second sentence\}}\\
    Answer: \\
    \bottomrule
 \end{tabular}
 \caption{Few-shot exemplars prompt for the WiC task.}\label{tab.wic.few-shot} 
\end{table}

\begin{table*}[!hbp] 
  \centering
 \footnotesize
  % \scriptsize
  % \tiny
   \begin{tabular}{
    m{0.96\textwidth}
    }
    \toprule
    Yes/No question-answering consists of a short passage and a Yes/No question about the passage. The questions are provided anonymously and unsolicited by users of the Google search engine, and afterwards paired with a paragraph from a Wikipedia article containing the answer. If there exists evidence in the passage that supports the facts in the question, the answer should be "Yes". If there exists evidence in the passage that denies the facts in the question, the answer should be "No".\\
    Your task is to read the passage and predict whether the answer to the question is "Yes" or "No".\\
    \\
    Passage: The Elder Scrolls Online -- As with other games in The Elder Scrolls series, the game is set on the continent of Tamriel. The events of the game occur a millennium before those of The Elder Scrolls V: Skyrim and around 800 years before The Elder Scrolls III: Morrowind and The Elder Scrolls IV: Oblivion. It has a broadly similar structure to Skyrim, with two separate conflicts progressing at the same time, one with the fate of the world in the balance, and one where the prize is supreme power on Tamriel. In The Elder Scrolls Online, the first struggle is against the Daedric Prince Molag Bal, who is attempting to meld the plane of Mundus with his realm of Coldharbour, and the second is to capture the vacant imperial throne, contested by three alliances of the mortal races. The player character has been sacrificed to Molag Bal, and Molag Bal has stolen their soul, the recovery of which is the primary game objective.\\
    Question: is elder scrolls online the same as skyrim\\
    Answer: No\\
    \\
    Passage: Good Samaritan law -- Good Samaritan laws offer legal protection to people who give reasonable assistance to those who are, or who they believe to be, injured, ill, in peril, or otherwise incapacitated. The protection is intended to reduce bystanders' hesitation to assist, for fear of being sued or prosecuted for unintentional injury or wrongful death. An example of such a law in common-law areas of Canada: a good Samaritan doctrine is a legal principle that prevents a rescuer who has voluntarily helped a victim in distress from being successfully sued for wrongdoing. Its purpose is to keep people from being reluctant to help a stranger in need for fear of legal repercussions should they make some mistake in treatment. By contrast, a duty to rescue law requires people to offer assistance and holds those who fail to do so liable.\\
    Question: do good samaritan laws protect those who help at an accident\\
    Answer: Yes\\
    \\
    Passage: Windows Movie Maker -- Windows Movie Maker (formerly known as Windows Live Movie Maker in Windows 7) is a discontinued video editing software by Microsoft. It is a part of Windows Essentials software suite and offers the ability to create and edit videos as well as to publish them on OneDrive, Facebook, Vimeo, YouTube, and Flickr.\\
    Question: is windows movie maker part of windows essentials\\
    Answer: Yes\\
    \\
    Passage: Epsom railway station -- Epsom railway station serves the town of Epsom in Surrey. It is located off Waterloo Road and is less than two minutes' walk from the High Street. It is not in the London Oyster card zone unlike Epsom Downs or Tattenham Corner stations. The station building was replaced in 2012/2013 with a new building with apartments above the station (see end of article).\\
    Question: can you use oyster card at epsom station\\
    Answer: No\\
    \\
    Passage: Da Vinci's Demons -- The series premiered in the United States on Starz on 12 April 2013, and its second season premiered on 22 March 2014. The series was renewed for a third season, which premiered on 24 October 2015. On 23 July 2015, Starz announced that the third season would be the show's last. However Goyer has left it open for a miniseries return.\\
    Question: will there be a season 4 of da vinci's demons\\
    Answer: No\\
    \\
    Passage: Powdered sugar -- Powdered sugar, also called confectioners' sugar, icing sugar, and icing cake, is a finely ground sugar produced by milling granulated sugar into a powdered state. It usually contains a small amount of anti-caking agent to prevent clumping and improve flow. Although most often produced in a factory, powdered sugar can also be made by processing ordinary granulated sugar in a coffee grinder, or by crushing it by hand in a mortar and pestle.\\
    Question: is confectionary sugar the same as powdered sugar\\
    Answer: Yes\\
         \end{tabular}
 \caption{Continued on the next page}
\end{table*} 

\begin{table*}[t]
 \ContinuedFloat
  \centering
 \footnotesize
  % \scriptsize
  % \tiny
   \begin{tabular}{
    m{0.96\textwidth}
    }
    % \toprule
    Passage: Federal judiciary of the United States -- The federal courts are composed of three levels of courts. The Supreme Court of the United States is the court of last resort. It is generally an appellate court that operates under discretionary review, which means that the Court can choose which cases to hear, by granting writs of certiorari. There is therefore generally no basic right of appeal that extends automatically all the way to the Supreme Court. In a few situations (like lawsuits between state governments or some cases between the federal government and a state) it sits as a court of original jurisdiction.\\
    Question: is the federal court the same as the supreme court\\
    Answer: No\\
    \\
    Passage: Bixby letter -- In the 1998 war film Saving Private Ryan, General George Marshall (played by Harve Presnell) reads the Bixby letter to his officers before giving the order to find and send home Private James Francis Ryan after Ryan's three brothers died in battle.\\
    Question: did abraham lincoln write the letter in saving private ryan\\
    Answer: Yes\\
    \\
    Passage: \textbf{\{passage\}}\\
    Question: \textbf{\{question\}}\\
    Answer: \\
    \bottomrule
 \end{tabular}
 \caption{Few-shot exemplars prompt for the BoolQ task.}\label{tab.boolq.few-shot} 
\end{table*}

% \clearpage
\newpage
\section{Few-shot CoT Prompts}\label{sec.appendix.d}

\begin{table*}[bh] 
  \centering
 \footnotesize
  % \scriptsize
  % \tiny
   \begin{tabular}{
    m{0.96\textwidth}
    }
    \toprule
    Given a search engine query, first, consider what the user could have in mind when they type in the query and allow for misspellings or other ambiguity, then classify the relevance of keyword to the query into one of the following categories: "Not bad", or "Bad". The definitions of the categories are\\
    "Not bad": the keyword is relevant to the user's search query. This can include: broader or narrower product selection, competitor or alternative products, accessories, products often purchased together and related topics as well as direct matches to the user's search.\\
    "Bad": the keyword is not relevant to the user's search query. There is no relationship between the query and keyword. This includes but is not limited to: incorrect/unrelated product, unrelated topic, wrong location when location is important, cannot be used in place of query product nor are they commonly purchased together, etc.\\
    Please predict whether the keyword is relevant to the query or not. The answer should be exact "Not bad" or "Bad".\\
    \\
    Query: google data studio sharepoint\\
    Keyword: sharepoint migration tool file share\\
    Answer: \textbf{\underline{The relevance of the keyword "sharepoint migration tool file share" to the search query "google data} \underline{studio sharepoint" is "Bad".} This keyword is not relevant to the user's search query as it pertains to a different tool or technology (SharePoint) that is not mentioned in the search query. The keyword focuses on a migration tool for SharePoint file sharing, while the search query is focused on Google Data Studio. There is no relationship between the two, so the keyword is not relevant to the user's search query.} \textit{Therefore, the relevance is "Bad".}\\
    \\
    Query: motorhomes sale\\
    Keyword: rv sale used class c\\
    Answer: \textbf{\underline{The relevance of keyword "rv sale used class c" is "Not bad".} The keyword "rv sale used class c" contains relevant information to the user's search query "motorhomes sale". It includes a specific type of motorhome, "class c", that is being sold used, which is a relevant product selection for the user's search. Additionally, the keyword "rv sale used class c" is a direct match to the user's search and is commonly purchased together with the products in the search query.} \textit{Therefore, the relevance is "Not bad".}\\
    \\
    Query: southern exposure seed exchange company\\
    Keyword: uk poppy seeds\\
    Answer: \textbf{\underline{The relevance of the keyword "uk poppy seeds" to the search query "southern exposure seed exchange} \underline{company" is "Not bad".} This is because the search query is about a seed exchange company, and the keyword "uk poppy seeds" is a specific type of seed that could be offered by such a company. Even though the company mentioned in the search query is based in the southern hemisphere, they may still offer seeds from other regions, including the UK, which could include poppy seeds.} \textit{Therefore, the relevance is "Not bad".}\\
    \\
    Query: nissan parts canada\\
    Keyword: purchase tires\\
    Answer: \textbf{\underline{The relevance of "purchase tires" to the query "nissan parts canada" is "Bad".} The keyword "purchase tires" is not directly related to the user's search for Nissan parts in Canada. Tires are not a part of a Nissan car and therefore are not relevant to the search for Nissan parts. Additionally, the keyword does not relate to a common purchase with Nissan parts, and it is not a substitute for the search query.} \textit{Therefore, the relevance is "Bad".}\\
    \\
    Query: \textbf{\{query\}}\\
    Keyword: \textbf{\{keyword\}}\\
    Answer: \\
    \bottomrule
 \end{tabular}
 \caption{Few-shot exemplars for full chain-of-thought prompt for the QK task. The bold text indicates the explanations  generated with the prompt in Table  \ref{tab.qk.explanation}.}\label{tab.qk.few-shot-cot} 
\end{table*}

\begin{table*}[!hbp] 
  \centering
 \footnotesize
  % \scriptsize
  % \tiny
   \begin{tabular}{
    m{0.96\textwidth}
    }
    \toprule
    Task description: The task is to identify if the occurrences of w in s1 and s2 correspond to the same meaning or not. The target word w in s1 and s2 is enclosed in quotation marks. There is no need to consider the tense and form of w. If w in s1 and s2 corresponds to the same meaning, output "true", otherwise output "false". \\
    \\
    w: "place"
    s1: Do you want to come over to my "place" later?\\
    s2: A political system with no "place" for the less prominent groups.\\
    Explanation: \textbf{The answer is "false". In the first sentence, "place" refers to a physical location or a person's home. In the second sentence, "place" refers to a position or status in a political system. Therefore, although the spelling and punctuation of "place" are the same in both sentences, the context and meaning of the word are different.} Therefore, the answer is "false".\\
    \\
    w: "hold"
    s1: The general ordered the colonel to "hold" his position at all costs.\\
    s2: "Hold" the taxi.\\
    Explanation: \textbf{The answer is "true". In both sentences, "hold" means to maintain or stay in a specific position or location. In the first sentence, the general orders the colonel to hold his position at all costs, meaning the colonel should not retreat or move from his current location. In the second sentence, someone is asking the taxi driver to hold, meaning to wait and stay in the current location until the person returns. Therefore, in both sentences, "hold" is used to convey the idea of staying in a particular position or location.} Therefore, the answer is "true".\\
    \\
    w: "summer"\\
    s1: We like to "summer" in the Mediterranean.\\
    s2: We "summered" in Kashmir.\\
    Explanation: \textbf{The answer is "true". In both sentences, "summer" refers to the activity of spending a summer vacation in a certain location. Although the word form is different ("summer" in the first sentence and "summered" in the second), the meaning is the same. The use of quotation marks in both sentences implies that "summer" is being used as a verb, rather than a season. Therefore, both sentences convey the idea of enjoying a summer vacation in different locations, making the meaning the same.} Therefore, the answer is "true".\\
    \\
    w: "approach"\\
    s1: "Approach" a task.\\
    s2: To "approach" the city.\\
    Explanation: \textbf{The answer is "false". In the first sentence, "Approach" is used as a verb and means to begin or undertake a task. In the second sentence, "approach" is also used as a verb but means to move towards or get closer to the city. Although the spelling and the quotation marks around the word are the same in both sentences, the context and the meaning of the word are different. Therefore, the occurrences of "approach" in the two sentences do not correspond to the same meaning.} Therefore, the answer is "false".\\
    \\
    w: "cover"\\
    s1: "Cover" her face with a handkerchief.\\
    s2: Count the cash in the drawer twice just to "cover" yourself.\\
    Explanation: \textbf{The answer is "false". The word "cover" has different meanings in the two sentences. In the first sentence, "cover" means to hide or conceal something. In contrast, in the second sentence, "cover" means to protect oneself from potential problems or accusations. The context and usage of the word "cover" in the two sentences are different, indicating that the two instances of "cover" do not correspond to the same meaning.} Therefore, the answer is "false".\\
    \\
    w: "head"\\
    s1: His horse won by a "head".\\
    s2: He is two "heads" taller than his little sister.\\
    Explanation: \textbf{The answer is "true". In both sentences, "head" is used as a unit of measurement. In the first sentence, it means that the speaker's horse won by a small margin, specifically the length of the horse's head. In the second sentence, it means that the person is taller than their little sister by a specific amount, namely two head lengths. Therefore, in both cases, "head" is used as a measure of distance or height, indicating that the two occurrences correspond to the same meaning.} Therefore, the answer is "true".\\
    \\
    w: "meet"\\
    s1: The company agrees to "meet" the cost of any repairs.\\
    s2: This proposal "meets" my requirements.\\
    Explanation: \textbf{The answer is "true". Although the verb "meet" is used in different senses in the two sentences, in both cases, it conveys the idea of satisfying a requirement or obligation. In the first sentence, the company agrees to pay for any repairs, which is a requirement or obligation. In the second sentence, the proposal fulfills or satisfies the speaker's requirements. Therefore, even though the two sentences use "meet" in different senses, they convey the same general idea of satisfying an obligation or requirement, making the answer "true".} Therefore, the answer is "true".\\
        % \bottomrule
 \end{tabular}
 \caption{Continued on the next page}
\end{table*} 

\begin{table*}
 \ContinuedFloat
  \centering
 \footnotesize
  % \scriptsize
  % \tiny
   \begin{tabular}{
    m{0.96\textwidth}
    }
    % \toprule
    w: "development"\\
    s1: The organism has reached a crucial stage in its "development".\\
    s2: Our news team brings you the latest "developments".\\
    Explanation: \textbf{The answer is "false". In the first sentence, "development" refers to the growth or progress of an organism. In contrast, in the second sentence, "developments" refer to the latest events or updates on a particular subject. The first sentence refers to a specific stage in the life of an organism, while the second sentence refers to recent events or news related to a particular topic. Therefore, the context and meaning of "development" in both sentences are different, and they cannot be considered the same.} Therefore, the answer is "false".\\
    \\
    w: \textbf{\{target word\}}\\
    s1: \textbf{\{first sentence\}}\\
    s2: \textbf{\{second sentence\}}\\
    Explanation: \\
    \bottomrule
 \end{tabular}
 \caption{Few-shot exemplars for full chain-of-thought prompt for the WiC task. The bold text indicates the explanations generated with the prompt in Table  \ref{tab.wic.1}.}\label{tab.wic.few-shot-cot} 
\end{table*}

\begin{table*}[!hbp] 
  \centering
 \footnotesize
  % \scriptsize
  % \tiny
   \begin{tabular}{
    m{0.96\textwidth}
    }
    \toprule
    Yes/No question-answering consists of a short passage and a Yes/No question about the passage. The questions are provided anonymously and unsolicited by users of the Google search engine, and afterwards paired with a paragraph from a Wikipedia article containing the answer. If there exists evidence in the passage that supports the facts in the question, the answer should be "Yes". If there exists evidence in the passage that denies the facts in the question, the answer should be "No".\\
    Your task is to read the passage and predict whether the answer to the question is "Yes" or "No".\\
    \\
    Passage: The Elder Scrolls Online -- As with other games in The Elder Scrolls series, the game is set on the continent of Tamriel. The events of the game occur a millennium before those of The Elder Scrolls V: Skyrim and around 800 years before The Elder Scrolls III: Morrowind and The Elder Scrolls IV: Oblivion. It has a broadly similar structure to Skyrim, with two separate conflicts progressing at the same time, one with the fate of the world in the balance, and one where the prize is supreme power on Tamriel. In The Elder Scrolls Online, the first struggle is against the Daedric Prince Molag Bal, who is attempting to meld the plane of Mundus with his realm of Coldharbour, and the second is to capture the vacant imperial throne, contested by three alliances of the mortal races. The player character has been sacrificed to Molag Bal, and Molag Bal has stolen their soul, the recovery of which is the primary game objective.\\
    Question: is elder scrolls online the same as skyrim\\
    Answer: \textbf{The answer is "No". Although The Elder Scrolls Online is set on the continent of Tamriel, like Skyrim and other games in The Elder Scrolls series, it takes place a millennium before Skyrim and around 800 years before Morrowind and Oblivion. While it has a similar structure to Skyrim, with two conflicts progressing at the same time, the primary struggle in The Elder Scrolls Online is against the Daedric Prince Molag Bal, who has stolen the player character's soul, and the second is to capture the vacant imperial throne. Therefore, The Elder Scrolls Online is not the same game as Skyrim.} Therefore, the answer is "No".\\
    \\
    Passage: Good Samaritan law -- Good Samaritan laws offer legal protection to people who give reasonable assistance to those who are, or who they believe to be, injured, ill, in peril, or otherwise incapacitated. The protection is intended to reduce bystanders' hesitation to assist, for fear of being sued or prosecuted for unintentional injury or wrongful death. An example of such a law in common-law areas of Canada: a good Samaritan doctrine is a legal principle that prevents a rescuer who has voluntarily helped a victim in distress from being successfully sued for wrongdoing. Its purpose is to keep people from being reluctant to help a stranger in need for fear of legal repercussions should they make some mistake in treatment. By contrast, a duty to rescue law requires people to offer assistance and holds those who fail to do so liable.\\
    Question: do good samaritan laws protect those who help at an accident\\
    Answer: \textbf{ The answer is "Yes". Good Samaritan laws offer legal protection to people who provide reasonable assistance to those who are injured, ill, in peril, or otherwise incapacitated. The purpose of these laws is to encourage bystanders to provide help without fear of being sued or prosecuted for unintentional injury or wrongful death. In Canada, for example, the good Samaritan doctrine is a legal principle that prevents a rescuer who has voluntarily helped a victim in distress from being successfully sued for wrongdoing. Therefore, individuals who provide reasonable assistance at an accident can be protected by good Samaritan laws.} Therefore, the answer is "Yes".\\
    \\
    Passage: Windows Movie Maker -- Windows Movie Maker (formerly known as Windows Live Movie Maker in Windows 7) is a discontinued video editing software by Microsoft. It is a part of Windows Essentials software suite and offers the ability to create and edit videos as well as to publish them on OneDrive, Facebook, Vimeo, YouTube, and Flickr.\\
    Question: is windows movie maker part of windows essentials\\
    Answer: \textbf{The answer to the question "is Windows Movie Maker part of Windows Essentials?" is Yes. As stated in the passage, Windows Movie Maker is a discontinued video editing software by Microsoft and was a part of the Windows Essentials software suite. Windows Essentials was a collection of free software programs from Microsoft, including Photo Gallery, Movie Maker, Mail, Writer, and OneDrive, among others. Therefore, Windows Movie Maker was one of the programs included in the Windows Essentials suite, which was available for download from Microsoft's website.} Therefore, the answer is "Yes".\\
            % \bottomrule
 \end{tabular}
 \caption{Continued on the next page}
\end{table*} 

\begin{table*}[!hbp] 
 \ContinuedFloat
  \centering
 \footnotesize
  % \scriptsize
  % \tiny
   \begin{tabular}{
    m{0.96\textwidth}
    }
    % \toprule
    Passage: Epsom railway station -- Epsom railway station serves the town of Epsom in Surrey. It is located off Waterloo Road and is less than two minutes' walk from the High Street. It is not in the London Oyster card zone unlike Epsom Downs or Tattenham Corner stations. The station building was replaced in 2012/2013 with a new building with apartments above the station (see end of article).\\
    Question: can you use oyster card at epsom station\\
    Answer: \textbf{The answer is "No". The passage explicitly states that Epsom railway station is not in the London Oyster card zone, unlike Epsom Downs or Tattenham Corner stations. Therefore, you cannot use Oyster card at Epsom railway station.} Therefore, the answer is "No".\\
    \\
    Passage: Da Vinci's Demons -- The series premiered in the United States on Starz on 12 April 2013, and its second season premiered on 22 March 2014. The series was renewed for a third season, which premiered on 24 October 2015. On 23 July 2015, Starz announced that the third season would be the show's last. However Goyer has left it open for a miniseries return.\\
    Question: will there be a season 4 of da vinci's demons\\
    Answer: \textbf{The answer is "No". The passage states that the third season of "Da Vinci's Demons" was the show's last and that the series creator, David S. Goyer, left open the possibility of a miniseries return. However, there is no indication that a fourth season of the show was planned or produced. Therefore, based on the information provided, it can be concluded that there will not be a season 4 of "Da Vinci's Demons".} Therefore, the answer is "No".\\
    \\
    Passage: Powdered sugar -- Powdered sugar, also called confectioners' sugar, icing sugar, and icing cake, is a finely ground sugar produced by milling granulated sugar into a powdered state. It usually contains a small amount of anti-caking agent to prevent clumping and improve flow. Although most often produced in a factory, powdered sugar can also be made by processing ordinary granulated sugar in a coffee grinder, or by crushing it by hand in a mortar and pestle.\\
    Question: is confectionary sugar the same as powdered sugar\\
    Answer: \textbf{The answer is "Yes". Confectioners' sugar is another name for powdered sugar, as mentioned in the passage. It is produced by milling granulated sugar into a fine powder, and usually contains a small amount of anti-caking agent to prevent clumping. Both terms, confectioners' sugar and powdered sugar, can be used interchangeably to refer to the same product.} Therefore, the answer is "Yes".\\
    \\
    Passage: Federal judiciary of the United States -- The federal courts are composed of three levels of courts. The Supreme Court of the United States is the court of last resort. It is generally an appellate court that operates under discretionary review, which means that the Court can choose which cases to hear, by granting writs of certiorari. There is therefore generally no basic right of appeal that extends automatically all the way to the Supreme Court. In a few situations (like lawsuits between state governments or some cases between the federal government and a state) it sits as a court of original jurisdiction.\\
    Question: is the federal court the same as the supreme court\\
    Answer: \textbf{The answer is "No". The federal court is not the same as the Supreme Court. The federal courts consist of three levels of courts, while the Supreme Court is the highest level court of the federal judiciary. The Supreme Court operates under discretionary review and generally only hears cases on appeal. It also has original jurisdiction in certain specific situations, but it is not the same as the federal court.} Therefore, the answer is "No".\\
    \\
    Passage: Batman \& Robin (film) -- With the box office success of Batman Forever in June 1995, Warner Bros. immediately commissioned a sequel. They hired director Joel Schumacher and writer Akiva Goldsman to reprise their duties the following August, and decided it was best to fast track production for a June 1997 target release date, which is a break from the usual 3-year gap between films. Schumacher wanted to homage both the broad camp style of the 1960s television series and the work of Dick Sprang. The storyline of Batman \& Robin was conceived by Schumacher and Goldsman during pre-production on A Time to Kill. Portions of Mr. Freeze's back-story were based on the Batman: The Animated Series episode ``Heart of Ice'', written by Paul Dini.\\
    Question: is batman and robin a sequel to batman forever\\
    Answer: \textbf{The answer is "Yes". Based on the information provided in the passage, Batman \& Robin is a sequel to Batman Forever. Warner Bros. commissioned the sequel after the box office success of Batman Forever, and hired director Joel Schumacher and writer Akiva Goldsman to continue their duties. The film was fast-tracked for a June 1997 release, with Schumacher and Goldsman conceiving the storyline during pre-production of another movie. Therefore, Batman \& Robin is a direct sequel to Batman Forever, released just two years later.} Therefore, the answer is "Yes".\\
    \\
    Passage: \textbf{\{passage\}}\\
    Question: \textbf{\{question\}}\\
    Answer: \\
    \bottomrule
 \end{tabular}
 \caption{Few-shot exemplars for full chain-of-thought prompt for the BoolQ task. The bold text indicates the explanations generated with the prompt in Table  \ref{tab.boolq.1}.}
 \label{tab.boolq.few-shot-cot} 
\end{table*}

\clearpage
% \newpage
\section{Prompts Used to Test the Stability}\label{sec.appendix.e}
We present the few-shot prompts p1, p2 and p3 in Tables \ref{tab.boolq.few-shot1}, \ref{tab.boolq.few-shot2} and \ref{tab.boolq.few-shot3}, respectively. 
The few-shot prompt p3 is obtained by swapping the order of the ``Question'' and ``Passage'' in Table \ref{tab.boolq.few-shot}. 
While few-shot prompts p1 and p2 have minor variations in their task description compared to p3, we have highlighted the differences in bold. 
The few-shot prompts, p1, p2, and p3, consist of the same demonstrated examples as the original prompt presented in Table \ref{tab.boolq.few-shot}.

We show the few-shot CoT prompts p1, p2 and p3 in Tables \ref{tab.boolq.few-shot-cot1}, \ref{tab.boolq.few-shot-cot2} and \ref{tab.boolq.few-shot-cot3}, respectively. 
The few-shot CoT prompt p3 is obtained by swapping the order of the ``Question'' and ``Passage'' in Table \ref{tab.boolq.few-shot-cot}. 
While few-shot CoT prompts p1 and p2 have minor variations in their task description compared to p3, we have highlighted the differences in bold. 
The few-shot CoT prompts, p1, p2, and p3, consist of the same demonstrated examples as the original prompt presented in Table \ref{tab.boolq.few-shot-cot}.

\begin{table}[!hbp] 
  \centering
 \footnotesize
  % \scriptsize
  % \tiny
   \begin{tabular}{
    m{0.46\textwidth}
    }
    \toprule
    \textbf{Your task is to read the passage and predict whether the answer to the question is "Yes" or "No".}\\
    \\
    \textbf{Question}: is elder scrolls online the same as skyrim\\
    \textbf{Passage}: The Elder Scrolls Online -- As with other games in The Elder Scrolls series, the game is set on the continent of Tamriel. The events of the game occur a millennium before those of The Elder Scrolls V: Skyrim and around 800 years before The Elder Scrolls III: Morrowind and The Elder Scrolls IV: Oblivion. It has a broadly similar structure to Skyrim, with two separate conflicts progressing at the same time, one with the fate of the world in the balance, and one where the prize is supreme power on Tamriel. In The Elder Scrolls Online, the first struggle is against the Daedric Prince Molag Bal, who is attempting to meld the plane of Mundus with his realm of Coldharbour, and the second is to capture the vacant imperial throne, contested by three alliances of the mortal races. The player character has been sacrificed to Molag Bal, and Molag Bal has stolen their soul, the recovery of which is the primary game objective.\\
    Answer: No\\
    \\
    \textbf{......} \\
    \bottomrule
 \end{tabular}
 \caption{Few-shot prompt p1 for the BoolQ task. 
 }
 \label{tab.boolq.few-shot1} 
\end{table}

\begin{table}[!hbp] 
  \centering
 \footnotesize
  % \scriptsize
  % \tiny
   \begin{tabular}{
    m{0.46\textwidth}
    }
    \toprule
    \textbf{Yes/No question-answering consists of a short passage and a Yes/No question about the passage. The questions are provided anonymously and unsolicited by users of the Google search engine, and afterwards paired with a paragraph from a Wikipedia article containing the answer.}\\
    \textbf{Your task is to read the passage and predict whether the answer to the question is "Yes" or "No".}\\
    \\
    \textbf{Question}: is elder scrolls online the same as skyrim\\
    \textbf{Passage}: The Elder Scrolls Online -- As with other games in The Elder Scrolls series, the game is set on the continent of Tamriel. The events of the game occur a millennium before those of The Elder Scrolls V: Skyrim and around 800 years before The Elder Scrolls III: Morrowind and The Elder Scrolls IV: Oblivion. It has a broadly similar structure to Skyrim, with two separate conflicts progressing at the same time, one with the fate of the world in the balance, and one where the prize is supreme power on Tamriel. In The Elder Scrolls Online, the first struggle is against the Daedric Prince Molag Bal, who is attempting to meld the plane of Mundus with his realm of Coldharbour, and the second is to capture the vacant imperial throne, contested by three alliances of the mortal races. The player character has been sacrificed to Molag Bal, and Molag Bal has stolen their soul, the recovery of which is the primary game objective.\\
    Answer: No\\
    \\
    \textbf{......} \\
    \bottomrule
 \end{tabular}
 \caption{Few-shot prompt p2 for the BoolQ task. 
 }
 \label{tab.boolq.few-shot2} 
\end{table}

\begin{table}[!hbp] 
  \centering
 \footnotesize
  % \scriptsize
  % \tiny
   \begin{tabular}{
    m{0.46\textwidth}
    }
    \toprule
    \textbf{Yes/No question-answering consists of a short passage and a Yes/No question about the passage. The questions are provided anonymously and unsolicited by users of the Google search engine, and afterwards paired with a paragraph from a Wikipedia article containing the answer. If there exists evidence in the passage that supports the facts in the question, the answer should be "Yes". If there exists evidence in the passage that denies the facts in the question, the answer should be "No".}\\
    \textbf{Your task is to read the passage and predict whether the answer to the question is "Yes" or "No".}\\
    \\
    \textbf{Question}: is elder scrolls online the same as skyrim\\
    \textbf{Passage}: The Elder Scrolls Online -- As with other games in The Elder Scrolls series, the game is set on the continent of Tamriel. The events of the game occur a millennium before those of The Elder Scrolls V: Skyrim and around 800 years before The Elder Scrolls III: Morrowind and The Elder Scrolls IV: Oblivion. It has a broadly similar structure to Skyrim, with two separate conflicts progressing at the same time, one with the fate of the world in the balance, and one where the prize is supreme power on Tamriel. In The Elder Scrolls Online, the first struggle is against the Daedric Prince Molag Bal, who is attempting to meld the plane of Mundus with his realm of Coldharbour, and the second is to capture the vacant imperial throne, contested by three alliances of the mortal races. The player character has been sacrificed to Molag Bal, and Molag Bal has stolen their soul, the recovery of which is the primary game objective.\\
    Answer: No\\
    \\
    \textbf{......} \\
    \bottomrule
 \end{tabular}
 \caption{Few-shot prompt p3 for the BoolQ task. 
 }
 \label{tab.boolq.few-shot3} 
\end{table}

\begin{table*}[!hbp] 
  \centering
 % \footnotesize
  \scriptsize
  % \tiny
   \begin{tabular}{
    m{0.98\textwidth}
    }
    \toprule
    \textbf{Your task is to read the passage and predict whether the answer to the question is "Yes" or "No".}\\
    \\
    \textbf{Question}: is elder scrolls online the same as skyrim\\
    \textbf{Passage}: The Elder Scrolls Online -- As with other games in The Elder Scrolls series, the game is set on the continent of Tamriel. The events of the game occur a millennium before those of The Elder Scrolls V: Skyrim and around 800 years before The Elder Scrolls III: Morrowind and The Elder Scrolls IV: Oblivion. It has a broadly similar structure to Skyrim, with two separate conflicts progressing at the same time, one with the fate of the world in the balance, and one where the prize is supreme power on Tamriel. In The Elder Scrolls Online, the first struggle is against the Daedric Prince Molag Bal, who is attempting to meld the plane of Mundus with his realm of Coldharbour, and the second is to capture the vacant imperial throne, contested by three alliances of the mortal races. The player character has been sacrificed to Molag Bal, and Molag Bal has stolen their soul, the recovery of which is the primary game objective.\\
    Answer: The answer is "No". Although The Elder Scrolls Online is set on the continent of Tamriel, like Skyrim and other games in The Elder Scrolls series, it takes place a millennium before Skyrim and around 800 years before Morrowind and Oblivion. While it has a similar structure to Skyrim, with two conflicts progressing at the same time, the primary struggle in The Elder Scrolls Online is against the Daedric Prince Molag Bal, who has stolen the player character's soul, and the second is to capture the vacant imperial throne. Therefore, The Elder Scrolls Online is not the same game as Skyrim. Therefore, the answer is "No".\\
    \textbf{......} \\
    \bottomrule
 \end{tabular}
 \caption{Few-shot CoT prompt p1 for the BoolQ task. 
 }
 \label{tab.boolq.few-shot-cot1} 
\end{table*}

\begin{table*}[!hbp] 
  \centering
 % \footnotesize
  \scriptsize
  % \tiny
   \begin{tabular}{
    m{0.98\textwidth}
    }
    \toprule
    \textbf{Yes/No question-answering consists of a short passage and a Yes/No question about the passage. The questions are provided anonymously and unsolicited by users of the Google search engine, and afterwards paired with a paragraph from a Wikipedia article containing the answer.}\\
    \textbf{Your task is to read the passage and predict whether the answer to the question is "Yes" or "No".}\\
    \\
    \textbf{Question}: is elder scrolls online the same as skyrim\\
    \textbf{Passage}: The Elder Scrolls Online -- As with other games in The Elder Scrolls series, the game is set on the continent of Tamriel. The events of the game occur a millennium before those of The Elder Scrolls V: Skyrim and around 800 years before The Elder Scrolls III: Morrowind and The Elder Scrolls IV: Oblivion. It has a broadly similar structure to Skyrim, with two separate conflicts progressing at the same time, one with the fate of the world in the balance, and one where the prize is supreme power on Tamriel. In The Elder Scrolls Online, the first struggle is against the Daedric Prince Molag Bal, who is attempting to meld the plane of Mundus with his realm of Coldharbour, and the second is to capture the vacant imperial throne, contested by three alliances of the mortal races. The player character has been sacrificed to Molag Bal, and Molag Bal has stolen their soul, the recovery of which is the primary game objective.\\
    Answer: The answer is "No". Although The Elder Scrolls Online is set on the continent of Tamriel, like Skyrim and other games in The Elder Scrolls series, it takes place a millennium before Skyrim and around 800 years before Morrowind and Oblivion. While it has a similar structure to Skyrim, with two conflicts progressing at the same time, the primary struggle in The Elder Scrolls Online is against the Daedric Prince Molag Bal, who has stolen the player character's soul, and the second is to capture the vacant imperial throne. Therefore, The Elder Scrolls Online is not the same game as Skyrim. Therefore, the answer is "No".\\
    \textbf{......} \\
    \bottomrule
 \end{tabular}
 \caption{Few-shot CoT prompt p2 for the BoolQ task. 
 }
 \label{tab.boolq.few-shot-cot2} 
\end{table*}

\begin{table*}[!hbp] 
  \centering
 % \footnotesize
  \scriptsize
  % \tiny
   \begin{tabular}{
    m{0.96\textwidth}
    }
    \toprule
    \textbf{Yes/No question-answering consists of a short passage and a Yes/No question about the passage. The questions are provided anonymously and unsolicited by users of the Google search engine, and afterwards paired with a paragraph from a Wikipedia article containing the answer. If there exists evidence in the passage that supports the facts in the question, the answer should be "Yes". If there exists evidence in the passage that denies the facts in the question, the answer should be "No".}\\
    \textbf{Your task is to read the passage and predict whether the answer to the question is "Yes" or "No".}\\
    \\
    \textbf{Question}: is elder scrolls online the same as skyrim\\
    \textbf{Passage}: The Elder Scrolls Online -- As with other games in The Elder Scrolls series, the game is set on the continent of Tamriel. The events of the game occur a millennium before those of The Elder Scrolls V: Skyrim and around 800 years before The Elder Scrolls III: Morrowind and The Elder Scrolls IV: Oblivion. It has a broadly similar structure to Skyrim, with two separate conflicts progressing at the same time, one with the fate of the world in the balance, and one where the prize is supreme power on Tamriel. In The Elder Scrolls Online, the first struggle is against the Daedric Prince Molag Bal, who is attempting to meld the plane of Mundus with his realm of Coldharbour, and the second is to capture the vacant imperial throne, contested by three alliances of the mortal races. The player character has been sacrificed to Molag Bal, and Molag Bal has stolen their soul, the recovery of which is the primary game objective.\\
    Answer: The answer is "No". Although The Elder Scrolls Online is set on the continent of Tamriel, like Skyrim and other games in The Elder Scrolls series, it takes place a millennium before Skyrim and around 800 years before Morrowind and Oblivion. While it has a similar structure to Skyrim, with two conflicts progressing at the same time, the primary struggle in The Elder Scrolls Online is against the Daedric Prince Molag Bal, who has stolen the player character's soul, and the second is to capture the vacant imperial throne. Therefore, The Elder Scrolls Online is not the same game as Skyrim. Therefore, the answer is "No".\\
    \textbf{......} \\
    \bottomrule
 \end{tabular}
 \caption{Few-shot CoT prompt p3 for the BoolQ task. 
 }
 \label{tab.boolq.few-shot-cot3} 
\end{table*}

\newpage
% \clearpage
\section{Prompts for Constructing the Conversation-based Information Retrieval Dataset}\label{sec.appendix.f}

\begin{table}[hbp!]
  \centering
 \footnotesize
  % \scriptsize
  % \tiny
   \begin{tabular}{
    m{0.48\textwidth}
    }
    \toprule
    Now you need to enrich the following text paragraphs with content that is relevant and factually consistent with the content of the text paragraphs.\\
    Passage: \textcolor{blue}{To determine the salary range percentile, you must first calculate the difference between the maximum and minimum salary figures. For example, if the salary range for a particular position is between \$45,000 and \$75,000, the difference between those two figures would be \$30,000.}\\
    \textcolor{red}{Once you have calculated the salary range difference, you can then use this information to determine the salary range percentile. This is a way to compare salaries of individuals in the same profession or industry, and can be helpful in negotiating salaries and assessing the competitiveness of compensation packages.}\\
    \textcolor{red}{The salary range percentile can be calculated by taking an individual's salary and comparing it to the range of salaries for the same position. For instance, if an individual's salary is \$60,000 and the salary range for their position is \$45,000 to \$75,000, their salary falls within the 50th percentile of the salary range.}\\
    \textcolor{red}{It's important to note that salary ranges can vary depending on a number of factors, including location, industry, experience, and education. Employers typically set salary ranges based on market research and benchmarking against similar positions in the same industry and geographic region.}\\
    \textcolor{red}{Additionally, some companies may have a more structured salary range system, while others may be more flexible and open to negotiation. It's important to research and understand the salary range for your position and industry to ensure that you are being compensated fairly and competitively.}\\
    % Passage: \textbf{\{passage\}}\\
    \bottomrule
 \end{tabular}
 \caption{Zero-shot prompt used to enrich text paragraphs (\textcolor{blue}{blue} = input; \textcolor{red}{red} = output).}\label{tab.enrich_text.zero-shot} 
\end{table}

\begin{table}[hbp!]
  \centering
 \footnotesize
  % \scriptsize
  % \tiny
   \begin{tabular}{
    m{0.48\textwidth}
    }
    \toprule
    I need you to simulate the scene of a conversation between a person and ChatGPT. P1 represents the person, who is responsible for asking questions based on the given text passage, and P2 represents the ChatGPT, who is responsible for answering the questions posed by P1. You need to answer query according to the following passage and form the first round of the dialogue:\\
    Query: \textcolor{blue}{how to determine salary range}\\
    Passage: \textcolor{blue}{Once you have calculated the salary range difference, you can then use this information to determine the salary range percentile. This is a way to compare salaries of individuals in the same profession or industry, and can be helpful in negotiating salaries and assessing the competitiveness of compensation packages.}\\
    \textcolor{blue}{The salary range percentile can be calculated by taking an individual's salary and comparing it to the range of salaries for the same position. For instance, if an individual's salary is \$60,000 and the salary range for their position is \$45,000 to \$75,000, their salary falls within the 50th percentile of the salary range.}\\
    \textcolor{blue}{It's important to note that salary ranges can vary depending on a number of factors, including location, industry, experience, and education. Employers typically set salary ranges based on market research and benchmarking against similar positions in the same industry and geographic region.}\\
    \textcolor{blue}{Additionally, some companies may have a more structured salary range system, while others may be more flexible and open to negotiation. It's important to research and understand the salary range for your position and industry to ensure that you are being compensated fairly and competitively.}\\
    % Query: \textbf{\{query\}}\\
    % Passage: \textbf{\{passage\}}\\
    Next, you need to continue several rounds of dialogue according to the first round of dialogue. Complete the above requirements step by step.\\

    \textcolor{red}{P1: How do you determine the salary range percentile?}\\
    \textcolor{red}{P2: You must first compute the difference between the highest and minimum wage figures in order to establish the salary range percentile.}\\
    \textcolor{red}{P1: Can you give me an example of this calculation?}\\
    \textcolor{red}{P2: Sure. For example, if the salary range for a particular position is between \$45,000 and \$75,000, the difference between those two figures would be \$30,000.}\\
    \textcolor{red}{P1: Is there anything else I need to do to determine the salary range?}\\
    \textcolor{red}{P2: Once you have calculated the difference between the maximum and minimum salary figures, you can use that information to generate the salary range percentile.}\\
    \textcolor{red}{P1: How do I generate the salary range percentile?}\\
    \textcolor{red}{P2: The salary range percentile is generated by comparing the salary range for a particular position to the salaries for similar positions in the same industry and geographic location.}\\
    \textcolor{red}{P1: Is the wage range percentile generated using a specific formula?}\\
    \textcolor{red}{P2: There is no specific formula for generating the salary range percentile, as it can vary depending on the company and industry. However, it typically involves researching salary data for similar positions and adjusting the salary range based on factors such as experience level, education, and geographic location.}\\
    \bottomrule
 \end{tabular}
 \caption{Zero-shot prompt for conversation generation (\textcolor{blue}{blue} = input; \textcolor{red}{red} = output).}\label{tab.conversation_generation.zero-shot} 
\end{table}

\begin{table*}[bh] 
  \centering
 % \footnotesize
  \scriptsize
  % \tiny
   \begin{tabular}{
    m{0.96\textwidth}
    }
    \toprule
    Given a conversation and passage, first, consider the relevance of the conversation and paragraph, explore the relevant content between the conversation and passage, and then classify the relevance between the conversation and passage into one of the following categories: "Relevant" or "Not Relevant Enough". The definitions of the categories are:\\
    "Not Relevant Enough": The conversation is not relevant enough to the passage, as a small portion of the content may have appeared in the passage, but most of the content in the conversation has not appeared in the passage or is not related to the passage. The content involved in the conversation may not be consistent with the facts described in the passage. For the specific details in the passage, the conversation did not involve relevant details.\\
    "Relevant": Passage and conversation are related. The topic of the conversation is consistent with the paragraph, and the facts mentioned in the conversation are consistent with the facts in the passage. The content of the conversation is mostly related to the paragraph, and can be appropriately expanded.\\
    Here are some examples:\\
    <Passage>\\
    To determine the salary range percentile, you must first calculate the difference between the maximum and minimum salary figures. For example, if the salary range for a particular position is between \$45,000 and \$75,000, the difference between those two figures would be \$30,000.\\
% \end{tabular}
%  \caption{Continued on the next page}
% \end{table*} 

% \begin{table*}[!hbp] 
%  \ContinuedFloat
%   \centering
%  \footnotesize
%   % \scriptsize
%   % \tiny
%    \begin{tabular}{
%     m{0.96\textwidth}
%     }
    <Conversation>\\
    P1: How do you determine the salary range percentile?\\
    P2: You must first compute the difference between the highest and minimum wage figures in order to establish the salary range percentile.\\
    P1: Can you give me an example of this calculation?\\
    P2: Sure. For example, if the salary range for a particular position is between $45,000 and $75,000, the difference between those two figures would be \$30,000.\\
    P1: Is there anything else I need to do to determine the salary range?\\
    P2: Once you have calculated the difference between the maximum and minimum salary figures, you can use that information to generate the salary range percentile.\\
    P1: How do I generate the salary range percentile?\\
    P2: The salary range percentile is generated by comparing the salary range for a particular position to the salaries for similar positions in the same industry and geographic location.\\
    P1: Is the wage range percentile generated using a specific formula?\\
    P2: There is no specific formula for generating the salary range percentile, as it can vary depending on the company and industry. However, it typically involves researching salary data for similar positions and adjusting the salary range based on factors such as experience level, education, and geographic location.\\
    <Answer>\\
    The conversation is discussing how to determine the salary range percentile, which is the topic of the passage. The conversation provides an example that is consistent with the information presented in the passage, and also expands on how to generate the salary range percentile by comparing it to salaries for similar positions. The information in the conversation is relevant and consistent with the information presented in the passage. Therefore, the relecance is "Relevant".\\
    <end>\\
    \\
    <Passage>\\
    An antigen is any molecule against which a specific immune response can be generated. This means that lymphocytes are activated: 1  B-cells are activated to produce antibodies and. 2  T-cells are activated to provide helper function and cytotoxic function.\\
    <Conversation>\\
    P1: What is the function of lymphocytes?\\
    P2: Lymphocytes are a type of white blood cell that are responsible for producing antibodies.\\
    P1: What do antibodies do?\\
    P2: Antibodies attach to the antigens on the surface of bacteria or viruses, which helps the immune system identify and destroy the pathogen.\\
    P1: Can you explain what antigens are?\\
    P2: Antigens are molecules that are found on the surface of bacteria or viruses. They are recognized by the immune system as foreign invaders, and the attachment of antibodies to these antigens helps to destroy the pathogen.\\
    P1: Are different antibodies needed for different types of antigens?\\
    P2: Yes, different organisms have different antigens, so a different antibody is needed to recognize each different type of antigen.\\
    P1: What happens when an antibody attaches to a virus?\\
    P2: When an antibody attaches to a virus, it can either destroy the pathogen or make it easier for a phagocyte to ingest and destroy it. This is an important part of the immune response to viral infections.\\
    <Answer>\\
    Although the conversation is related to the immune system and antibodies, it does not provide any additional information about the functions of B-cells or T-cells as described in the passage. The conversation also does not mention anything about the specific immune response generated by antigens, which is the main topic of the passage. While the conversation provides general information about the immune system and antibodies, it does not expand on the content of the passage or provide any additional relevant details. Therefore, the relecance is "Not Relevant Enough".\\
    <end>\\
    \\
    Please predict whether the conversation is relevant to the passage or not. The answer should be exact "Not Relevant Enough" or "Relevant".\\
    \\
    <Passage>\\
    \textbf{\{passage\}}\\
    <Conversation>\\
    \textbf{\{conversation\}}\\
    \bottomrule
 \end{tabular}
 \caption{Few-shot chain-of-thought prompt used to filter out irrelevant conversations.}\label{tab.data_filter.few-shot-cot} 
\end{table*} 

\clearpage
\section{Details on Human Evaluation}\label{sec.appendix.g}
For the purpose of human evaluation, we begin by presenting annotators with a multi-turn conversation accompanied by a paired passage. Their task involves carefully reading both the conversation and passage, ensuring a comprehensive grasp of the main topics and any significant details. Subsequently, they are required to assess the fluency of the conversation, as well as its relevance and consistency with the provided passage.
\subsection{Fluency}
To evaluate the fluency of the generated conversation, annotators should answer the first question:\\
How fluent do you think the conversation is? 

Following previous study \cite{he-yiu-2022-controllable}, annotators need to score the fluency of the conversation on a 5-point Likert scale from 1 to 5, based on the following rules:\\
    1: The conversation cannot be understood and all segments are not fluent.\\
    2: The conversation cannot be understood, but some segments are fluent.\\
    3: The conversation can be understood to some extent, but with many grammatical errors.\\
    4: The conversation can be understood with several grammatical errors.\\
    5: The conversation is extremely fluent without any grammatical errors.\\

\subsection{Relevance}
To assess the relevance between the generated conversation and the paired passage, graders need to consider the second question:\\
Q2: How relevant do you think the conversation is to the given passage? 

Specifically, graders need to score the relevance between the conversation and the given passage on a 3-point Likert scale from 1 to 3:\\
    1 (Irrelevant): Any topic discussed in the conversation is completely unrelated to the given passage.\\
    2 (Not Relevant Enough): Few topics discussed in the conversation are related to the given passage. \\
    3 (Relevant): Most topics discussed in the conversation are related to the given passage.\\
    
\subsection{Consistency}
As for consistency, graders should answer the following question: \\
Q3: How consistent do you think the conversation is to the given passage? 

To be concrete, graders need to score the consistency between the conversation and the given passage on a 3-point Likert scale from 1 to 3:\\
    1: Any fact mentioned in the conversation does not appear in the given passage.\\
    2: Few facts mentioned in the conversation are supported by the facts in the given passage.\\
    3: Most facts mentioned in the conversation are consistent with the facts in the passage. \\

We show the human evaluation results in Table \ref{table.ConIR.human2}. 
\begin{table}
\footnotesize
  \centering
    \begin{tabular}{
     m{0.06\textwidth}<{\centering}|
     m{0.1\textwidth}<{\centering}
     m{0.1\textwidth}<{\centering}
     m{0.1\textwidth}<{\centering}
     }
    \hline
    \textbf{Score} & \textbf{Fluency}  &\textbf{Relevance} &\textbf{Consistency} \\
    \hline 
    1 & 0 & 5& 16\\
    2 & 0 & 132& 144\\
    3 & 0& 163& 140\\
    4 & 3 & -& -\\
    5 & 297 & -& -\\
    \hline
    Average & 4.99 & 2.53& 2.41\\
    \hline

    \hline
  \end{tabular}
  \caption{Human evaluation results on ConIR. The first five rows display the frequency distribution of each annotation score. The last row represents the average score of the annotations.
  }\label{table.ConIR.human2}
\end{table}

\end{document}